\documentclass{article}

\usepackage{microtype}
\usepackage{graphicx}
\usepackage{subcaption}
\usepackage{booktabs} % for professional tables

\usepackage{hyperref}

\usepackage[preprint]{icml2026}

\usepackage{amsmath}
\usepackage{amssymb}
\usepackage{mathtools}
\usepackage{amsthm}

\usepackage{booktabs}
\usepackage{multirow}
\usepackage{graphicx}
\usepackage[table]{xcolor}
\usepackage{tcolorbox}
\usepackage[capitalize,noabbrev]{cleveref}
\usepackage{soul} % 用于下划线
\usepackage{listings} % 用于代码高亮
\theoremstyle{plain}

\theoremstyle{definition}

\theoremstyle{remark}

\usepackage{soul} % 用于下划线
\usepackage{listings} % 用于代码高亮

\usepackage{tikz}
\usepackage{xcolor}
\usepackage{listings}

\usepackage{enumitem}

\definecolor{myblue}{RGB}{30, 144, 255}    % 关键字颜色
\definecolor{mygreen}{RGB}{34, 139, 34}    % 字符串颜色
\definecolor{mybg}{RGB}{252, 252, 255}     % 文本框背景色
\definecolor{myframe}{RGB}{60, 60, 60}     % 边框颜色
\definecolor{highlighter}{RGB}{255, 240, 180} % 高亮色
\definecolor{codebg}{RGB}{250, 250, 250}
\definecolor{framecolor}{RGB}{50, 50, 50}
\definecolor{keywordcolor}{RGB}{0, 0, 128} % 深蓝
\definecolor{stringcolor}{RGB}{0, 100, 0}  % 深绿
\definecolor{instructioncolor}{RGB}{70, 70, 70}
\definecolor{codebg}{RGB}{250, 250, 252}       % 浅灰背景
\definecolor{framecolor}{RGB}{60, 60, 60}      % 深灰边框
\definecolor{headercolor}{RGB}{40, 40, 40}     % 标题背景
\definecolor{labelcolor}{RGB}{0, 50, 150}      % Q/A 标签颜色
\definecolor{highlight}{RGB}{200, 0, 0}        % 强调色

\definecolor{bg_color}{RGB}{252, 252, 252}     % 整体背景灰白
\definecolor{border_color}{RGB}{40, 40, 60}    % 深色边框
\definecolor{header_color}{RGB}{50, 60, 90}    % 标题蓝灰
\definecolor{entity_color}{RGB}{0, 100, 200}   % 实体高亮色
\definecolor{pheno_color}{RGB}{200, 0, 100}    % 表型高亮色
\definecolor{diag_color}{RGB}{0, 100, 0}       % 诊断高亮色
\usetikzlibrary{shapes, positioning, shadows, arrows.meta, calc}
\definecolor{bglight}{RGB}{245, 245, 245}  % 浅灰背景

\usepackage[textsize=tiny]{todonotes}
\definecolor{lightblue}{RGB}{173, 216, 230} 

\icmltitlerunning{PathReasoner-R1: Instilling Structured Reasoning into Computational Pathology Vision-Language Model}

\begin{document}

\twocolumn[
  \icmltitle{PathReasoner-R1: Instilling Structured Reasoning into Pathology Vision-Language Model via Knowledge-Guided Policy Optimization}



  \icmlsetsymbol{equal}{*}

  \begin{icmlauthorlist}
    \icmlauthor{Songhan Jiang}{equal,hit}
    \icmlauthor{Fengchun Liu}{equal,hit}
    \icmlauthor{Ziyue Wang}{mic,nus}
    \icmlauthor{Linghan Cai}{hit,tud}
    \icmlauthor{Yongbing Zhang}{hit}
    % \icmlauthor{}{sch}
    % \icmlauthor{}{sch}
    % \icmlauthor{}{sch}
  \end{icmlauthorlist}
  
  \icmlaffiliation{hit}{Harbin Institute of Technology (Shenzhen)}
  \icmlaffiliation{mic}{Microsoft Research}
  \icmlaffiliation{nus}{National University of Singapore}
  \icmlaffiliation{tud}{Technical University of Dresden}
  \icmlcorrespondingauthor{Linghan Cai}{cailh@stu.hit.edu.cn}
  \icmlcorrespondingauthor{Yongbing Zhang}{ybzhang08@hit.edu.cn}
    

  \vskip 0.3in
]



\printAffiliationsAndNotice{\icmlEqualContribution}  % no special notice (required even if empty)

\begin{abstract}
Vision-Language Models (VLMs) are advancing computational pathology with superior visual understanding capabilities. However, current systems often reduce diagnosis to directly output conclusions without verifiable evidence-linked reasoning, which severely limits clinical trust and hinders expert error rectification. To address these barriers, we construct PathReasoner, the first large-scale dataset of whole-slide image (WSI) reasoning. Unlike previous work reliant on unverified distillation, we develop a rigorous knowledge-guided generation pipeline. By leveraging medical knowledge graphs, we explicitly align structured pathological findings and clinical reasoning with diagnoses, generating over 20K high-quality instructional samples. Based on the database, we propose PathReasoner-R1, which synergizes trajectory-masked supervised fine-tuning with reasoning-oriented reinforcement learning to instill structured chain-of-thought capabilities. To ensure medical rigor, we engineer a knowledge-aware multi-granular reward function incorporating an Entity Reward mechanism strictly aligned with knowledge graphs. This effectively guides the model to optimize for logical consistency rather than mere outcome matching, thereby enhancing robustness. Extensive experiments demonstrate that PathReasoner-R1 achieves state-of-the-art performance on both PathReasoner and public benchmarks across various image scales, equipping pathology models with transparent, clinically grounded reasoning capabilities. Dataset and code are available at \href{https://github.com/cyclexfy/PathReasoner-R1}{https://github.com/cyclexfy/PathReasoner-R1}.
\end{abstract}

\section{Introduction}
% 背景：主流方法仍停留在“特征记忆”和“模式匹配”阶段
% 问题：缺乏知识，数据缺乏
% 挑战：WSI大，图像推理的弊端，文本思维链条数据集
% 本文方案
% 贡献
% Large Vision-Language Models (LVLMs)~\cite{Qwen3-VL,InternVL,Gemini} have exhibited exceptional proficiency in open-world visual understanding and question answering in recent years. In computational pathology, researchers have begun pairing pre-trained Large Language Models (LLMs) with pathology images to build interactive diagnostic assistants. However, most existing systems still frame diagnosis as memorization and imitation tasks, e.g., report generation and visual question answering (VQA)~\cite{WSI-LLaVA,WSI-VQA,PathGen,SlideChat}. As illustrated in Figure~\ref{fig:motiv}, they tend to predict a final diagnosis without providing supporting evidence or reasoning, which hinders clinical adoption and makes error rectification difficult. 

\begin{figure}[t]
  \centering
  \includegraphics[width=\columnwidth]{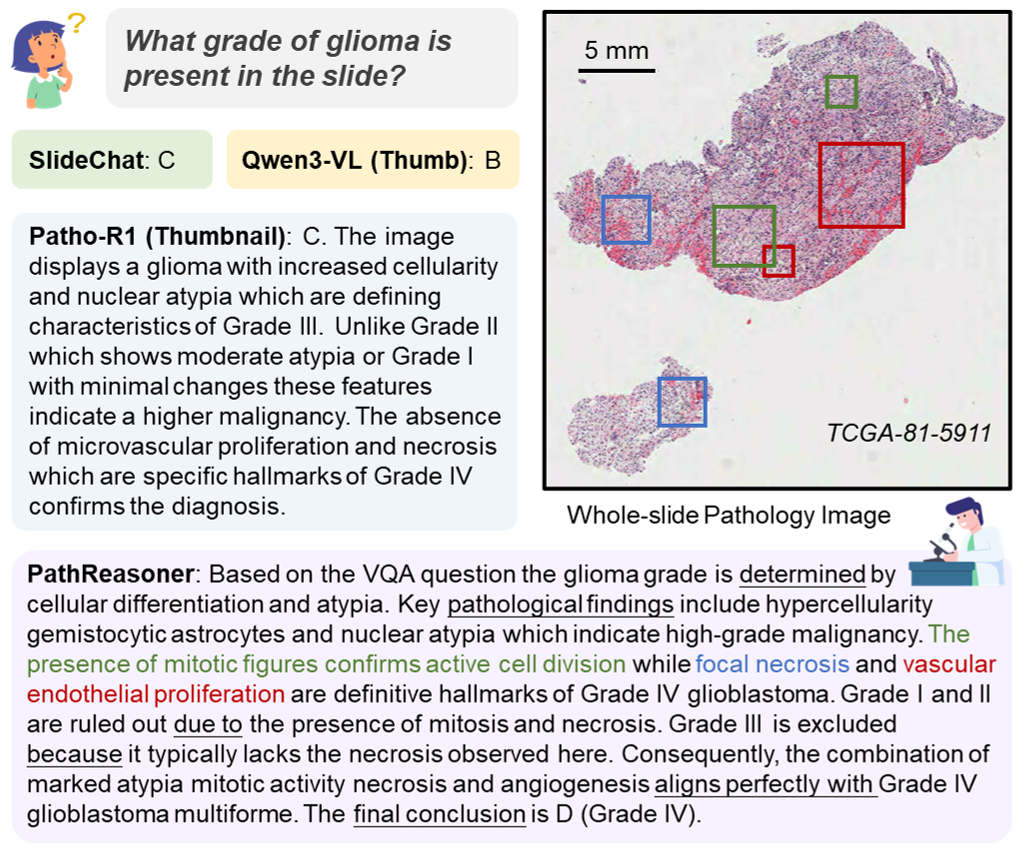}
  \caption{Comparisons of mainstream vision-language models in computational pathology. Existing models like SlideChat and Qwen3-VL perform direct diagnosis, while Patho-R1 generates superficial reasoning. In contrast, our PathReasoner-R1 employs medically grounded step-by-step reasoning, explicitly linking visual evidence to the diagnosis. Text colors correspond to the bounding boxes, and underlines highlight the logical flow.}
  \label{fig:motiv}
\end{figure}

\begin{figure*}[t]
  \centering
  \includegraphics[width=\textwidth]{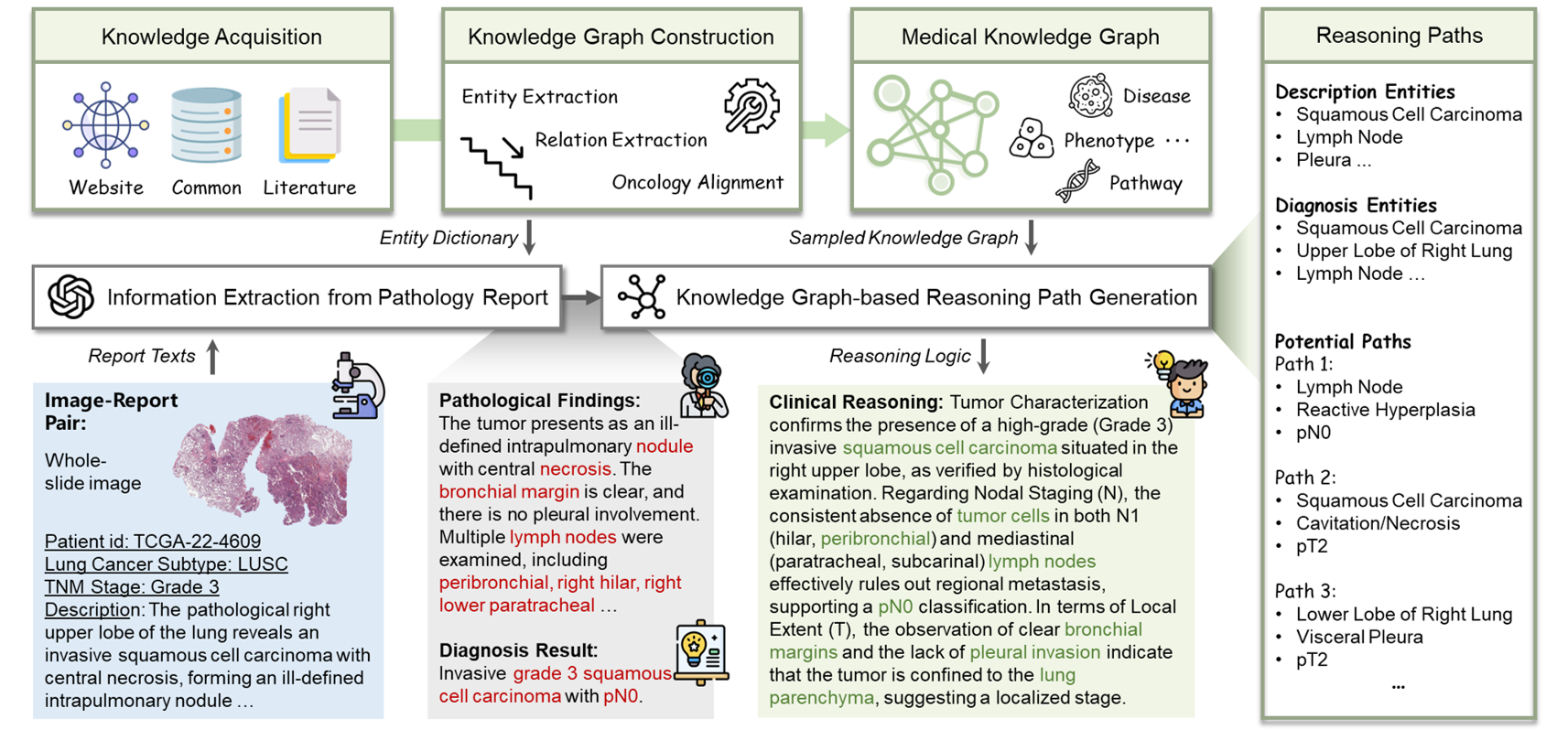}
  \caption{Overview of the PathReasoner construction pipeline. The framework transforms unstructured reports into structured CoT annotations through three key stages: constructing a medical knowledge graph from public platforms, aligning entities extracted from WSI pathology reports with graph nodes, and generating explicit reasoning paths that logically link visual findings to the final diagnosis.}
  \label{fig:data_construction}
\end{figure*}

The integration of Vision-Language Models (VLMs) into computational pathology (CPath) is establishing a new standard for interactive diagnostic assistants \cite{Qwen3-VL,lyu2025wsi,PathChat,cpathagent,wang2025medagent}. While recent CPath VLMs \cite{HistoGPT,WSI-LLaVA,SlideChat,Quilt-LLaVA,PathGen} have demonstrated proficiency in visual question answering (VQA) and image captioning, a fundamental reasoning gap remains evident, as illustrated in Figure~\ref{fig:motiv}. Current architectures predominantly formulate diagnosis as the direct prediction of diagnostic conclusions. This formulation often yields opaque predictions or hallucinates rationales without reliable evidential support, a limitation observed even in preliminary reasoning attempts like Patho-R1 \cite{Patho-R1}. Unlike human pathologists who strictly derive diagnosis conclusions through a structured chain of morphological evidence, such as identifying cellular atypia to rule out mimics, these models fail to provide explicit intermediate logic. This absence of transparent, evidence-based reasoning limits clinical interpretability and significantly impedes experts' ability to rectify model errors in high-stakes decision-making.

% Meanwhile, the broader VLLM community is shifting from pattern recognition to deliberate, multi-step reasoning, employing strategies like Chain-of-Thought (CoT) combined with reinforcement learning (RL) \cite{DeepSeek-R1}. However, in computational pathology, this transition is obstructed by a fundamental data bottleneck: the scarcity of large-scale supervision that records intermediate diagnostic findings and explicitly links them to slide evidence. Although recent initiatives \cite{TeamPath, Patho-R1} attempt to bridge this divide, they predominantly rely on synthetic CoT traces distilled from generic models (e.g., GPT-4o~\cite{GPT-4}), which often lack domain-specific medical precision. Compounding this limitation are practical constraints that force reliance on patch- or thumbnail-based processing. Such fragmentation disrupts the global context, preventing models from capturing the holistic visual evidence essential for diagnostic reasoning.
While the large language model (LLM) community is transitioning to deliberate reasoning via reinforcement learning (RL) \cite{DeepSeek-R1, yu2025medresearcher}, the CPath field faces a dual barrier. Existing attempts are stifled not only by a severe data bottleneck, specifically the scarcity of WSI-level chain-of-thought (CoT) annotations, but also by the lack of pathology-aligned supervision mechanisms \cite{TeamPath, pathology-cot}. For instance, models like Patho-R1 \cite{Patho-R1} are constrained to isolated ROIs, while WSI-level attempts such as SmartPath-R1 \cite{SmartPath} struggle under sparse, outcome-based reward signals. Without explicit alignment between structured pathological findings and diagnostic conclusions, RL algorithms fail to optimize the intermediate reasoning process. Consequently, even reasoning-oriented models often revert to hallucination or shortcut learning, producing rationales that are structurally plausible but medically superficial. This underscores that advanced training paradigms require a synergy of high-fidelity CoT data and granular, knowledge-aware reward functions to unlock their full potential.

To dismantle these barriers, we introduce PathReasoner, a framework centered on the first large-scale WSI-level reasoning dataset designed to align visual evidence with stepwise clinical logic. Unlike previous datasets reliant solely on black-box distillation \cite{SlideChat, WSI-LLaVA, Quilt-LLaVA}, PathReasoner is constructed through a rigorous knowledge-guided generation pipeline. We leverage medical knowledge graphs (KGs) (e.g., PrimeKG \cite{PrimeKG} and PathoGraph \cite{pathograph}) to inject verifiable clinical relationships into the synthesis process, ensuring that every step in the CoT, from identifying pathological findings to deducing the diagnosis, is grounded in established medical facts. This pipeline allows us to scale up high-quality annotations to over 20K samples, organizing each into a coherent structure comprising findings, reasoning, and diagnosis that mirrors the authentic diagnostic workflow of human pathologists.

% Building on \textit{PathReasoner}, we propose \textbf{\textit{PathReasoner-R1}}, a novel training framework designed to internalize these reasoning capabilities into pathology VLLMs. Our method employs a two-stage paradigm. First, we conduct a cold-start using SFT to establish basic behavioral patterns, applying strategic trajectory masking to enforce focused supervision on the CoT content.
% This is followed by a reasoning-oriented RL stage aimed at refining logical robustness. To address the inherent instability of RL, we engineer a multi-granular reward function comprising format, semantic, and entity rewards. Notably, the entity reward is strictly aligned with the KGs used in our dataset construction, thereby guaranteeing the validity of the optimization signals. Crucially, beyond simple accuracy, our reward model explicitly evaluates logical consistency—verifying whether the generated Pathological Findings provide sufficient evidence to support the subsequent Clinical Reasoning. This synergy between high-fidelity data and rigorous process supervision enables us to overcome common RL performance bottlenecks, unlocking substantial improvements in both diagnostic accuracy and reasoning reliability.
Building on the foundational reasoning data, we propose PathReasoner-R1. Instead of utilizing generic post-training methods, we design a knowledge-guided training paradigm tailored for CPath that synergizes trajectory-masked supervised fine-tuning (SFT) with criteria-aligned reinforcement learning. To ensure the RL process optimizes for medical truth rather than just plausibility, we engineer a knowledge-aware multi-granular reward function. This mechanism uniquely incorporates an Entity Reward that is strictly aligned with medical knowledge, effectively guiding the model's policy to follow correct reasoning paths. PathReasoner-R1 thus represents a shift from imitation-based learning to genuine, clinically grounded diagnostic reasoning. In summary, main contributions are threefold:
\begin{itemize}
\item We construct PathReasoner, the first large-scale CoT dataset for WSI analysis to date, which bridges the reasoning gap by explicitly aligning visual evidence with stepwise, knowledge-grounded clinical logic.
\item We propose PathReasoner-R1, a knowledge-guided reasoning framework that ensures methodological consistency from data construction to policy optimization. By synergizing trajectory-masked SFT with criteria-aligned RL, we create a pipeline where model updates are strictly governed by medical Entity Rewards. This enables the autonomous emergence of clinically grounded reasoning capabilities, optimizing for logical validity beyond surface-level label matching.
\item Extensive experiments demonstrate that our method achieves state-of-the-art performance on both internal and external benchmarks across multiple image scales. Crucially, PathReasoner-R1 provides transparent, verifiable reasoning trajectories, marking a significant step toward trustworthy AI in computational pathology.
\end{itemize}

\section{Related Work}
\paragraph{Vision-Language Models in CPath.}
% VLMs have revolutionized computational pathology in recent years. Multimodal diagnostic models such as SlideChat~\cite{SlideChat} HistoGPT~\cite{HistoGPT}, MUSK~\cite{MUSK}, and WSI-LLaVA~\cite{WSI-LLaVA}, have demonstrated remarkable capabilities in automated report generation and visual question answering, directly mapping medical imagery to diagnostic textual conclusions. Furthermore, interactive models like PathChat~\cite{PathChat} have advanced the field by multi-turn dialogues functioning as AI copilots to assist human experts.
% However, these systems remain fundamentally confined to the supervised fine-tuning paradigm. Consequently, they function primarily through rote memorization or pattern matching, failing to ignite the intrinsic reasoning and cognitive capabilities.
% While general-domain LLMs have evolved to perform complex deduction via CoT and RL, pathology-specific attempts like Patho-R1~\cite{Patho-R1} and TeamPath~\cite{TeamPath} face significant hurdles. Not only are they severely constrained by input resolution limits (e.g., relying on patches or thumbnails), but they also suffer from a lack of high-quality CoT datasets. This data scarcity severely limits the efficacy of RL, preventing it from fully eliciting the models' deep reasoning capabilities.
VLMs have significantly advanced CPath, with frameworks like SlideChat~\cite{SlideChat}, WSI-LLaVA~\cite{WSI-LLaVA}, and PathFLIP~\cite{pathflip} demonstrating remarkable performance in VQA and image captioning. While interactive agents such as SlideSeek~\cite{PathChat} have further enabled multi-turn copilot functionality, these systems remain largely confined to supervised fine-tuning. Consequently, they tend to rely on superficial statistical correlations rather than rigorous clinical logic. Although the general LLM community has successfully implemented CoT and RL for complex deduction, pathology-specific adaptations, such as Patho-R1~\cite{Patho-R1} and TeamPath~\cite{TeamPath}, face distinct challenges. These attempts are hampered by dual bottlenecks: input-resolution constraints (e.g., reliance on thumbnails or ROIs) that sacrifice global context, and a severe scarcity of high-quality CoT data, which collectively constrain the model's capacity to transcend simple instruction-following and develop autonomous, deep diagnostic reasoning.

\begin{figure}[t]
  \centering
  \includegraphics[width=0.45\textwidth]{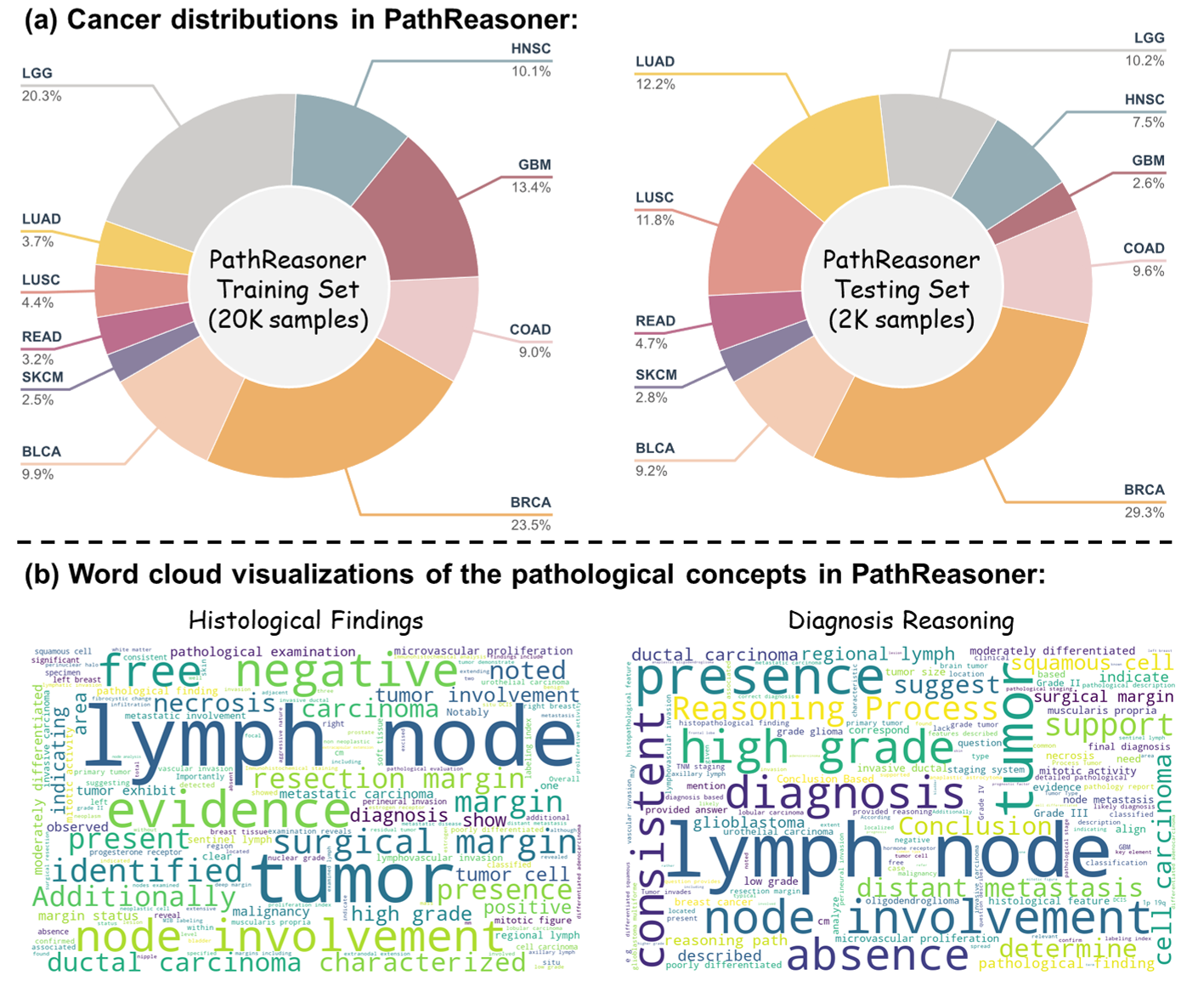}
  \caption{Statistical overview of PathReasoner. (a) Data distribution across 10 cancer types and various anatomical sites. (b) Diversity of pathological concepts covered in the dataset.
  }
  \label{fig:data_distribution}
\end{figure}

\paragraph{Instruction Tuning Datasets in CPath.}
High-quality data serves as the cornerstone of VLM capabilities, yet current benchmarks exhibit critical gaps in reasoning depth and knowledge grounding. Traditional ROI-level datasets, such as PathVQA~\cite{PathVQA}, PathGen~\cite{PathGen}, and PathMMU~\cite{PathMMU}, primarily focus on direct classification or QA, mapping images to labels without intermediate evidence. Although recent comprehensive benchmarks like MedXpertQA~\cite{Medxpertqa} incorporate pathology subsets, they remain restricted to non-WSI images (e.g., ROIs) and are designed strictly for evaluation rather than model training. Even recent gigapixel-scale benchmarks like WSI-LLaVA~\cite{WSI-LLaVA} largely retain this direct mapping paradigm, omitting explicit reasoning traces. While emerging initiatives like Patho-R1~\cite{Patho-R1} and SmartPath-R1~\cite{SmartPath} aim to incorporate reasoning processes, they rely heavily on distillation from general-purpose LLMs rather than on verified medical sources. Consequently, these datasets often yield rationales that are structurally plausible but medically superficial, failing to capture the granular, step-by-step deductive logic required for rigorous clinical diagnosis. By contrast, we present PathReasoner, a large-scale WSI reasoning dataset constructed via a rigorous knowledge-guided pipeline that explicitly aligns visual evidence with verifiable clinical logic derived from medical knowledge graphs.

% Data serves as the cornerstone of model capabilities, yet existing open-source datasets in computational pathology suffer from critical limitations in both logical depth and contextual scale.
% Early benchmarks, such as PathVQA~\cite{PathVQA} and PathGen~\cite{PathGen}, primarily focus on direct question answering or classification, mapping images to labels without providing intermediate evidence. Similarly, while WSI-LLaVA~\cite{WSI-LLaVA} and WSI-VQA~\cite{WSI-VQA} have pioneered gigapixel-scale understanding, they largely retain this direct mapping paradigm, lacking explicit reasoning traces.
% More recently, initiatives like Patho-R1~\cite{Patho-R1}, TeamPath~\cite{TeamPath}, and SmartPath~\cite{SmartPath} have attempted to augment datasets with thinking processes, ranging from agent trajectories~\cite{Pathology-cot} to WSI-level CoT. However, a fundamental flaw persists: these datasets rely heavily on distillation from general-purpose LLMs rather than expert-curated medical knowledge. Consequently, they often lack granular, step-by-step deductive logic and fail to capture the nuanced applicability of domain-specific medical knowledge. This reliance results in reasoning chains that are structurally plausible but medically superficial, yielding a dataset landscape that fails to simultaneously embody rigorous clinical accuracy and the intrinsic reasoning attributes required by advanced VLMs.

\begin{figure*}[t]
  \centering
  \includegraphics[width=\textwidth]{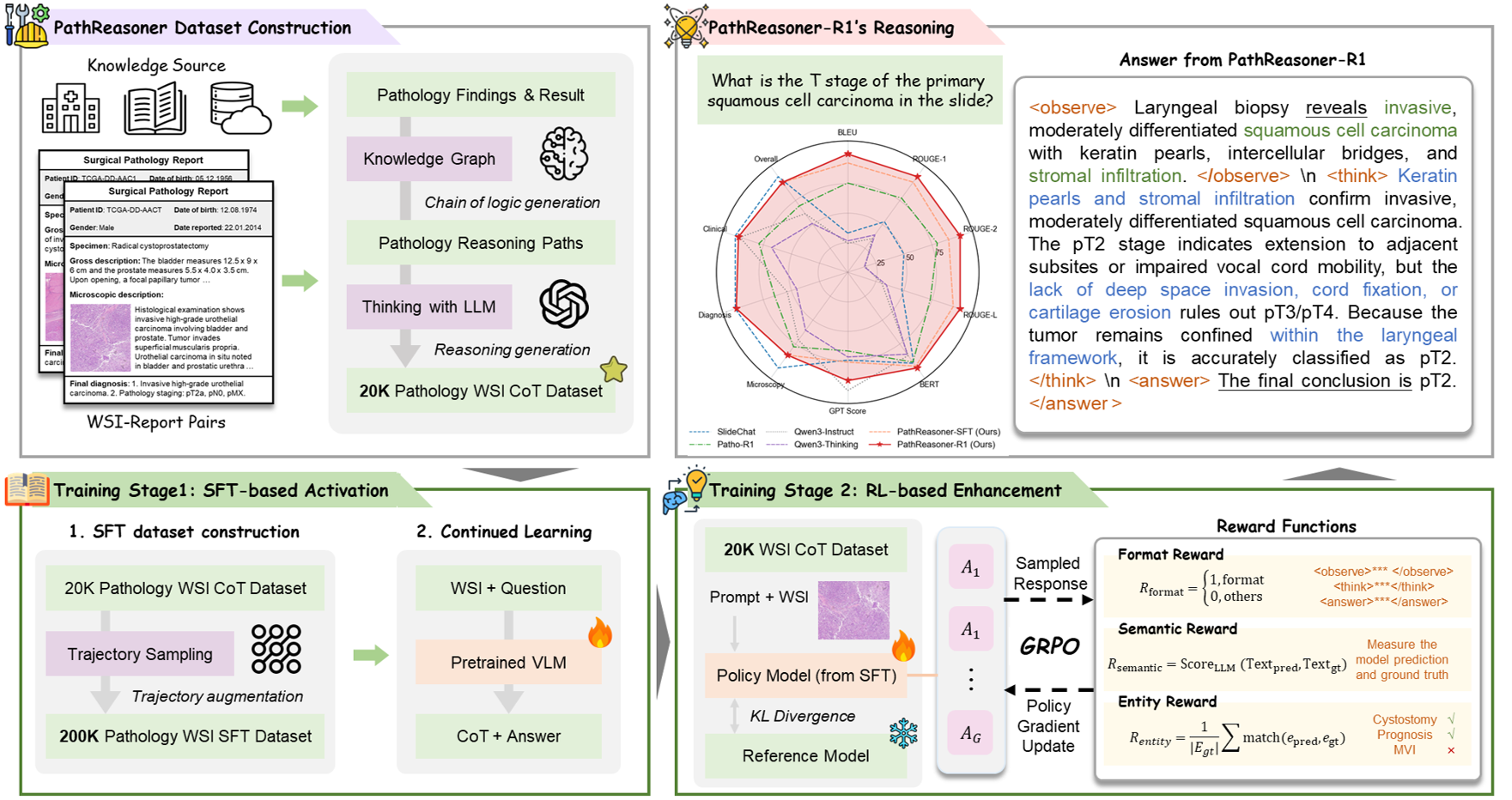}
  \caption{Overview of the PathReasoner-R1 framework. Building upon the PathReasoner dataset, the framework implements a two-stage post-training process for SlideChat: SFT-based reasoning activation and RL-based reasoning enhancement. This pipeline sequentially generates the initial policy model, PathReasoner-SFT-7B, and the final model, PathReasoner-R1-7B. The resulting model is optimized for open-ended VQA tasks, delivering well-organized outputs with superior reasoning capabilities.}
  \label{fig:overview}
\end{figure*}

\section{PathReasoner Construction}
To mitigate the scarcity of explicit reasoning datasets in CPath, we introduce \textbf{PathReasoner}, a large-scale, pathology reasoning dataset designed to enhance the capabilities of VLMs. We leverage original slide captions and diagnostic reports from TCGA as our foundational data, and we use medical/pathology knowledge graphs (KGs) to inject domain-specific knowledge. As illustrated in Figure \ref{fig:data_construction}, our construction pipeline consists of knowledge-guided generation followed by rigorous quality filtering. In the following, we present the construction details; additional descriptions are provided in Appendix~\ref{sec:construction}.

\noindent \textbf{Stage 1: KG Construction and Path Retrieval.}
To establish a structured reasoning foundation, we firstly construct a unified KG $\mathcal{G}$ by integrating PrimeKG~\cite{PrimeKG} with the PathoGraph ontology~\cite{pathograph}. 
% While PrimeKG provides broad biomedical context, we utilize PathoGraph's tripartite structure, comprising Entity, Phenotype, and Diagnosis graphs, to model the specific hierarchical composition of tissues (e.g., \texttt{hasComponent}) and the diagnostic logic. 
% To establish a structured reasoning foundation, we utilize \textbf{PrimeKG}~\cite{PrimeKG} as the macro-scale backbone, providing established relationships between diseases, genes, and clinical phenotypes. To resolve the lack of fine-grained visual concepts in PrimeKG, we alihttps://translate.google.com.hk/?sl=auto&tl=zh-CN&op=translategn it with the \textbf{PathoGraph ontology}~\cite{pathograph}.
Specifically, we map the ``Diagnosis" nodes in PathoGraph to ``Disease" nodes in PrimeKG. This fusion creates a continuous semantic pathway from micro-scale histological entities (e.g., \texttt{Physical\_Entity}, \texttt{Phenotype} from PathoGraph) to macro-scale clinical insights.
We leverage the TCGA dataset to instantiate distinct paths within $\mathcal{G}$. Diagnostic reports associated with WSIs are processed to extract structured evidence. We employ GPT-4o~\cite{GPT-4} for context-aware Named Entity Recognition (NER). Identified findings (e.g., ``nuclear atypia") are dynamically mapped to the corresponding $\{\hat{e}_i^{Q}\}_{i\in[n]}$ nodes (i.e., \texttt{Phenotype} or \texttt{Physical\_Entity}) in $\mathcal{G}$, serving as the starting points for graph-based reasoning.
Then, ground-truth answers are aligned with \texttt{Diagnosis} nodes (e.g., \texttt{Final\_Diagnosis}) to form end nodes $\{\hat{e}_i^{A}\}_{i\in[m]}$. Finally, we identify the shortest paths between entity anchors and these end nodes, prioritizing edges that encode diagnostic logic (e.g., \texttt{hasSupportEvidence}, \texttt{hasContradictEvidence}). This strategy reconstructs the \texttt{DiagnosisProcess}, capturing the direct causal chain from visual phenotypes to clinical outcomes. Appendices \ref{subsec:kg} and \ref{subsec:entity} show further details.

\noindent \textbf{Stage 2: Logic-Driven CoT Distillation.} Leveraging the paths extracted in Stage 1, we orchestrate GPT-4o to synthesize slide-level QA pairs via a knowledge-constrained generation strategy. By adopting these paths as the foundational reasoning backbone, we prompt the model to articulate the graph-encoded \texttt{DiagnosisProcess} into structured text. Specifically, the CoT generation is constrained to explicitly reference identified \texttt{PhysicalEntities} and \texttt{Phenotypes} as \texttt{SupportEvidence} before deriving the final diagnosis.
This distillation process ensures that GPT-4o’s reasoning is strictly grounded in medical facts, resulting in a dataset characterized by rigorous clinical logic.

% Using the KG nodes and paths extracted in Stage 1, we prompt GPT-4o to synthesize slide-level QA pairs. 
% Each answer contains an explicit, step-by-step reasoning trace guiding the VLMs from an under-specified query to the correct diagnosis.
% PathReasoner is the first slide-level image reasoning dataset that includes structured pathology findings, clinical reasoning, and complete CoT reasoning steps. This process effectively distills the reasoning capability of GPT-4o into our dataset.

\noindent \textbf{Stage 3: High-Quality CoT Filtering.}
To ensure clinical relevance and visual dependency, we formally define each sample as a triplet $\mathcal{T}=(Q, A, C)$ and apply a filtering protocol:
(1) Logical consistency check: We verify the internal coherence between the reasoning chain $C$ and the final answer $A$, discarding samples where the conclusion in $C$ contradicts $A$. (2) Visual dependency verification: To eliminate text-only biases, we employ GPT-4o in a blind setting to predict $A$ given only $Q$. Samples are discarded if the question leaks the answer without requiring visual evidence or reasoning. (3) Reasoning sufficiency validation: We assess whether the reasoning chain provides sufficient information to derive the ground truth. GPT-4o is prompted to infer the answer solely based on $C$; the triplet is retained only if the inferred answer aligns with $A$.
% \noindent \textbf{Stage 3: High-Quality CoT Filtering.}
% To ensure clinical relevance and image dependency, we apply a rigorous three-stage filtering process to each triplet $(Q, A, R)$:
% (1) Answer Verification: We verify the consistency between the reasoning chain $R$ and the final answer $A$, retaining only consistent samples.
% (2) Text Bias Elimination: To guarantee visual dependency, we discard samples where an LLM can correctly answer $Q$ using only textual context.
% (3) CoT Quality Verification: We validate if the reasoning chain is sufficient to derive the ground truth by prompting an LLM to extract the answer from $R$. We retain the chain only if the extracted answer matches $A$.

\noindent \textbf{Pathreasoner Attributes.} PathReasoner establishes a comprehensive benchmark with 22,153 samples, strategically partitioned into 20,153 for training and 2,000 for testing. The dataset covers 10 major cancer types, providing broad clinical coverage across diverse anatomical sites. This balanced representation, illustrated in Figure \ref{fig:data_distribution}(a), facilitates the learning of generalized pathological features rather than overfitting to specific organ characteristics. Beyond raw diagnostics, the textual data is structured into two distinct components: histopathology findings and clinical reasoning in Figure \ref{fig:data_distribution}(b). The former is densely populated with fine-grained morphological descriptors for rich visual-semantic grounding, while the latter features high-frequency causal keywords. Building on this structured data, the PathReasoner benchmark (testing set) is designed as an open-ended evaluation framework that covers multiple dimensions, including morphological descriptions, clinical diagnoses, and treatment plans. These comprehensive annotations make the diagnostic process transparent and verifiable, transforming the visual-question answering task from simple label prediction into interpretable deep inference.

% We construct the \textbf{PathReasoner} dataset by leveraging paired Whole Slide Images (WSIs) and their corresponding diagnostic reports from the TCGA \cite{TCGA} database. To ensure the diversity and robustness of the reasoning chains, we curated a large-scale collection encompassing $N$ major cancer types (e.g., LUAD, BRCA, COAD, etc.).
% Figures \ref{fig:data_distribution}(a) and (b) illustrate the distribution of cancer types across the training and test sets, respectively. The balanced representation of various anatomical sites ensures that the model can learn generalized pathological features rather than overfitting to specific organ characteristics.
% To further investigate the linguistic patterns and the depth of the reasoning process, we visualize the word frequency statistics for both \textit{histopathology findings} and \textit{clinical reasoning} components via word clouds in Figures \ref{fig:data_distribution}(c) and (d).
% First, the \textit{histopathology findings} (Figure c) are densely populated with fine-grained morphological descriptors such as ... which provides a rich visual-semantic grounding.
% Second, and more importantly, the \textit{clinical reasoning} part (Figure d) reveals a high frequency of causal and inferential keywords such as ...

\section{Methodology}
To fully unlock the potential of PathReasoner and equip CPath VLMs with rigorous clinical logic, we introduce \textbf{PathReasoner-R1}, a post-training framework built upon the WSI-level VLM, SlideChat \cite{SlideChat}. As shown in Figure~\ref{fig:overview}, the pipeline consists of two phases: (1) SFT-based reasoning activation, which leverages CoT data enriched with trajectory augmentation to establish the model's domain-specific logical foundation; and (2) RL-based reasoning enhancement, which employs Group Relative Policy Optimization (GRPO) \cite{GRPO} with tailored rewards to further elevate the VLM's reasoning capabilities.

\subsection{SFT-based Reasoning Activation} To enhance model generalization and mitigate the scarcity of process-oriented data, we introduce a trajectory augmentation strategy.
Rather than training exclusively on full sequences, we truncate reasoning chains at random intermediate steps. Specifically, given a reasoning chain $R=[s_1, s_2, \dots, s_L]$ sampled from the original dataset $\mathcal{D}$, we construct an augmented dataset $\mathcal{D}_{\text{aug}}$ as follows:
\begin{equation}
    \mathcal{D}_{\text{aug}} = \left\{ (x, q, \underbrace{s_{1:m-1}}_{\text{Context}}, \underbrace{s_{m:L}, a}_{\text{Target}}) \right\}_{m=1}^{L},
\end{equation}
where $s_{1:m-1}$ represents the visible context, and $s_{m:L}$ followed by $a$ is the target continuation.
By creating $L$ variations for each reasoning chain, this mechanism effectively scales our training corpus to 200K samples, enabling the model to robustly learn autoregressive logic recovery rather than simple pattern memorization. Appendix~\ref{subsec:trajectory} provides concrete examples.

Formally, for a sampled trajectory starting at index $m$, let $y$ denote the token sequence of the target segment $(s_{m:L}, a)$.
The training objective maximizes the likelihood of generating the target segment $y$:
\begin{equation} 
\mathcal{L}_{\text{SFT}} = -\mathbb{E}_{(x, q, \text{ctx}, y) \sim \mathcal{D}_{\text{aug}}}\sum_{t=1}^{|y|} \log \pi_\theta(y_{t} \mid x, q, \text{ctx}, y_{<t}), 
\end{equation}
where $\pi_\theta$ denotes the policy model's token distribution, and $\text{ctx} = {s_{1:m-1}}$ represents the visible context. This approach provides a rigorous reasoning activation foundation for the subsequent reinforcement learning phase.
% \begin{equation}
%     \mathcal{L}_{\text{SFT}} = -\mathbb{E}_{(x, q, y) \sim \mathcal{D}_{\text{aug}}} \sum_{t=m}^{T} \log \pi_\theta(y_{t} \mid x, q, y_{<t}) ,
% \end{equation}
% \begin{equation}
%     \mathcal{L}_{\text{SFT}} = -\mathbb{E}_{(x, q, y) \sim \mathcal{D}_{\text{aug}}}  \sum_{t \in \mathcal{T}_{\text{target}}} \log \pi_\theta(y_{t} \mid x, q, y_{<t}) 
% \end{equation}

\begin{table*}[t]
  \caption{Performance comparison on the PathReasoner benchmark. Metrics to the left of the thick vertical line evaluate answer quality; to the right, metrics evaluate chain-of-thought quality. 
  A-Score measures the semantic alignment of the reasoning chain with the ground truth; Q-Score evaluates the intrinsic logical coherence and step-wise quality of the reasoning process.
  The best and second-best results are highlighted in \textbf{bold} and \underline{underlined}, respectively. ``T'' and ``S'' denote Thumbnail and Slide inputs.
  } 
  \centering
  \renewcommand{\arraystretch}{1.15} % 调高，使表格更紧凑
  \scriptsize                        % 减小字号
  \setlength{\tabcolsep}{8pt}        % 减小列边距 (如果还宽，可以改为 2pt)
  \begin{tabular}{l @{\hspace{4pt}} c cccccc cc}
  \toprule
  \toprule
  \multirow{2}{*}{\textbf{Method}} & \multirow{2}{*}{} & \multicolumn{6}{c}{\textbf{Answer Quality Evaluation}} & \multicolumn{2}{c}{\textbf{Reasoning Quality Evaluation}} \\ 
  \cmidrule(lr){3-8} \cmidrule(lr){9-10}
   & & BLEU & ROUGE-1 & ROUGE-2 & ROUGE-L & BERT & LLM Score & A-Score & Q-Score \\ 
  \midrule
  \multicolumn{10}{l}{\cellcolor{gray!10}\textbf{Non-reasoning models}} \\
  Qwen2.5-VL-8B-Instruct    & T & 0.036 & 0.072 & 0.008 & 0.053 & 0.667 & 2.057 & - & - \\
  Qwen3-VL-8B-Instruct      & T & 0.059 & 0.110 & 0.020 & 0.084 & 0.678 & 2.120 & - & - \\
  LLaVA-Med-7B              & T & 0.123 & 0.195 & 0.042 & 0.143 & 0.687 & 1.083 & - & - \\
  HuatuoGPT-Vision-7B       & T & 0.076 & 0.145 & 0.039 & 0.115 & 0.699 & 1.214 & - & - \\
  MedGemma-4B-IT            & T & 0.058 & 0.110 & 0.031 & 0.088 & 0.673 & 1.280 & - & - \\
  Quilt-LLaVA-7B            & T & 0.061 & 0.109 & 0.019 & 0.087 & 0.670 & 1.940 & - & - \\
  SlideChat-7B (Baseline)   & S & 0.080 & 0.165 & 0.064 & 0.133 & 0.741 & 2.094 & - & - \\
  WSI-LLaVA-7B              & S & 0.100 & 0.164 & 0.027 & 0.112 & 0.714 & 1.675 & - & - \\
  \midrule \midrule
  \multicolumn{10}{l}{\cellcolor{gray!10}\textbf{Models with reasoning ability}} \\
  Qwen3-VL-8B-Thinking & T & 0.064 & 0.122 & 0.019 & 0.095 & 0.667 & 2.019 & 1.550 & 2.000 \\
  InternVL3.5-8B       & T & 0.043 & 0.092 & 0.021 & 0.081 & 0.693 & 2.028 & 1.853 & 3.310 \\
  MedVLThinker-7B      & T & 0.045 & 0.018 & 0.002 & 0.016 & 0.687 & 1.263 & 1.607 & 3.797 \\
  Patho-R1-7B          & T & 0.182 & 0.242 & 0.102 & 0.203 & 0.711 & 1.876 & 1.733 & 4.333 \\
  \rowcolor{blue!8}
  PathReasoner-SFT-7B & S & \underline{0.223} & \underline{0.294} & \underline{0.115} & \underline{0.259} & \underline{0.764} & \underline{2.143} & \underline{2.357} & \underline{4.598} \\
  \rowcolor{blue!11}
  PathReasoner-R1-7B  & S & \textbf{0.241} & \textbf{0.312} & \textbf{0.128} & \textbf{0.276} & \textbf{0.779} & \textbf{2.583} & \textbf{2.543} & \textbf{4.873} \\ \bottomrule \bottomrule
  \end{tabular}
  \label{tab:ours}
\end{table*}

\begin{table*}[t]
  \caption{Performance comparisons on the whole-slide image visual-question answering benchmarks. The best performances are in \textbf{bold}, the second-best performances are \underline{underlined}. ``T'' and ``S'' indicate the Thumbnail and Slide inputs, respectively.}
  \centering
  \renewcommand{\arraystretch}{1.10}
  \scriptsize
  \setlength{\tabcolsep}{7pt} 
  \begin{tabular}{lc cccc ccc}
  \toprule \toprule
  \textbf{Method} & & \multicolumn{4}{c}{\textbf{SlideBench-BCNB}} & \textbf{SlideBench-TCGA} & \textbf{WSI-VQA} & \textbf{CPTAC} \\
  \cmidrule(lr){3-6} \cmidrule(lr){7-7} \cmidrule(lr){8-8} \cmidrule(lr){9-9}
  & & Tumor type & Grading & Subtype & Average & Accuracy & Accuracy & Accuracy \\
  \midrule
  \rowcolor{gray!11}
  \multicolumn{9}{l}{\textbf{Non-reasoning models}} \\ 
  Qwen2.5-VL-8B-Instruct    & T & 58.13 & 27.11 & 16.07 & 34.06 & 41.48 & 33.43 & 32.92 \\ 
  Qwen3-VL-8B-Instruct      & T & 55.01 & 27.65 & 16.92 & 33.43 & 46.16 & 39.92 & 42.50 \\
  LLaVA-Med-7B              & T & 31.10 & 39.20 & 10.30 & 26.33 & 26.27 & 26.89 & 18.75 \\
  HuatuoGPT-Vision-7B       & T & 75.43 & 44.82 & 21.93 & 47.50 & 45.89 & 43.24 & 23.75 \\
  MedGemma-4B-IT            & T & 56.33 & 28.94 & 19.19 & 35.08 & 41.55 & 41.60 & 35.00 \\
  Quilt-LLaVA-7B               & T & 47.35 & 18.57 & 25.43 & 30.97 & 28.72 & 38.14 & 53.33 \\
  SlideChat-7B (Baseline)                 & S & \underline{90.20} & 23.10 & 17.50 & 43.60 & \textbf{75.36} & \underline{54.83} & 48.75 \\
  WSI-LLaVA-7B                 & S & \textbf{90.50} & \underline{46.30} & \underline{29.20} & \underline{55.30} & 60.20 & 42.05 & \underline{72.10} \\
  \midrule \midrule
  \rowcolor{gray!11}
  \multicolumn{9}{l}{\textbf{Models with reasoning ability}} \\
  Qwen3-VL-8B-Thinking  & T & 53.69 & 26.35 & 14.74 & 31.16 & 38.49 & 38.73 & 27.50 \\
  InternVL3.5-8B        & T & 48.87 & 44.38 & 24.48 & 39.02 & 49.82 & 39.62 & 33.33 \\
  MedVLThinker-7B       & T & 52.08 & 37.90 & 19.47 & 36.42 & 46.98 & 49.13 & 27.08 \\
  Patho-R1-7B           & T & 27.13 & 39.42 & 27.13 & 31.43 & 52.34 & 44.28 & 32.50 \\
  \rowcolor{blue!8}
  PathReasoner-SFT-7B   & S & 88.17 & 45.07 & 23.82 & 53.91 & 67.98 & 46.36 & 69.22 \\
  \rowcolor{blue!11}
  PathReasoner-R1-7B & S 
  & \underline{90.20} \tiny \textcolor[HTML]{D62627}{(-0.30)}
  & \textbf{52.23} \tiny \textcolor[HTML]{D62627}{(+5.93)}
  & \textbf{32.42} \tiny \textcolor[HTML]{D62627}{(+3.22)}
  & \textbf{57.68} \tiny \textcolor[HTML]{D62627}{(+2.38)}
  & \underline{74.68} \tiny \textcolor[HTML]{D62627}{(-0.68)}
  & \textbf{55.90} \tiny \textcolor[HTML]{D62627}{(+1.07)}
  & \textbf{74.95} \textcolor[HTML]{D62627}{(+2.85)} \\
  \bottomrule \bottomrule
  \end{tabular}%
  \label{tab:slide}
\end{table*}

\begin{table}[t]
\caption{Comparison of multi-modal large language models on the ROI-level benchmark PathMMU (full testing set). The best performances are in \textbf{bold}, the second-best performances are \underline{underlined}.}
\centering
\renewcommand{\arraystretch}{1.10}
\scriptsize 
\setlength{\tabcolsep}{1.8pt} % 调小列间距以适应单栏
\begin{tabular}{lcccccc}
\toprule
\toprule
\multirow{2}{*}{\textbf{Method}} & \multicolumn{6}{c}{\textbf{PathMMU-test}} \\
\cmidrule(lr){2-7}
 & Atlas & EduCont & PathCLS & PubMed & Social & Average \\
\midrule
\multicolumn{7}{l}{\cellcolor{gray!10}\textbf{Non-reasoning models}} \\
Qwen3-VL-8B-Instruct    & 21.40 & 21.51 & 7.72 & 21.31 & 20.54 & 18.59 \\
LLaVA-Med-7B            & 21.65 & 21.27 & 12.01 & 27.77 & 21.25 & 23.40 \\
HuatuoGPT-Vision-7B     & 58.07 & 54.72 & 36.64 & 61.36 & 59.37 & 54.59 \\
MedGemma-4B-IT          & 37.50 & 26.90 & 7.56 & 26.75 & 29.60 & 24.94 \\
Quilt-LLaVA-7B     & 41.43 & 36.72 & 14.71 & 34.80 & 35.29 & 32.02 \\
SlideChat-7B (Baseline) & 52.40 & 47.12 & 32.55 & 49.30 & 46.88 & 45.65 \\
WSI-LLaVA-7B            & 49.50 & 45.00 & 29.80 & 48.30 & 45.70 & 43.82 \\
\midrule
\midrule
\multicolumn{7}{l}{\cellcolor{gray!10}\textbf{Models with reasoning ability}} \\
Qwen3-VL-8B-Thinking    & 41.18 & 43.20 & 24.82 & 42.77 & 39.67 & 38.67 \\
InternVL3.5-8B       & 54.07 & 50.80 & 39.09 & 54.04 & 53.32 & 50.38 \\
MedVLThinker-7B      & 51.81 & 45.45 & 30.76 & 48.15 & 46.10 & 44.23 \\
Patho-R1-7B          & \textbf{75.34} & \textbf{66.43} & \underline{45.40} & \underline{66.06} & \textbf{67.93} & \underline{63.37} \\
\rowcolor{blue!8}
PathReasoner-SFT-7B & 62.56 & 61.82 & 44.72 & 60.46 & 60.23 & 57.96 \\
\rowcolor{blue!11}
PathReasoner-R1-7B & \underline{72.88} & \underline{65.78} & \textbf{50.03} & \textbf{66.17} & \underline{64.69} & \textbf{63.91} \\
\bottomrule
\bottomrule
\end{tabular}
\label{tab:patch}
\vspace{-1.0em}
\end{table}

% \subsection{RL-based Reasoning Enhancement}
% To further maximize reasoning reliability, we employ GRPO \cite{GRPO}, which optimizes the policy $\pi_\theta$ without a value network by estimating the baseline directly from group outcomes. For a given query $(x,q)$, we sample a group of $G$ outputs $\{a_i\}_{i=1}^G$ from the old policy $\pi_{\text{old}}$ and optimize the following objective:
% \begin{equation}
% \begin{split}
%     \mathcal{L}_{\text{GRPO}} = & -\mathbb{E}_{q \sim \mathcal{D}, \{a_i\} \sim \pi_{\text{old}}} \bigg[ \frac{1}{G}\sum_{i=1}^{G} \min \Big( r_i(\theta) A_i, \\
%     & \text{clip}\big(r_i(\theta), 1-\epsilon, 1+\epsilon\big) A_i \Big) \bigg] + \gamma \mathbb{D}_{KL}(\pi_\theta \| \pi_{\text{ref}})
% \end{split}
% \end{equation}
% where $\epsilon, \gamma$ are the clipping bound and KL regularization coefficient, respectively.
% The advantage $A_i$ is computed via group normalization to reduce variance: $A_i = (R(a_i) - \bar{R}) / \sigma_R$, where $\bar{R}$ and $\sigma_R$ denote the mean and standard deviation of rewards within the group.
% The optimization is guided by a knowledge-aware multi-granular reward function: $R(a_i) = R_{\text{format}}(a_i) + R_{\text{semantic}}(a_i) + \alpha R_{\text{entity}}(a_i)$.
%%
\subsection{RL-based Reasoning Enhancement}
To further maximize reasoning reliability, we employ GRPO \cite{GRPO}, which optimizes the policy $\pi_\theta$ by estimating from group outcomes. 
For a given query $(x,q)$, we sample a group of $G$ outputs $\{a_i\}_{i=1}^G$ from the old policy $\pi_{\text{old}}$.
The optimization objective is formulated as:
\begin{equation}
\begin{split}
    \mathcal{L}_{\text{GRPO}} = & -\mathbb{E}_{q \sim \mathcal{D}, \{a_i\} \sim \pi_{\text{old}}} \bigg[ \frac{1}{G}\sum_{i=1}^{G} \min \Big( r_i(\theta) A_i, \\
    & \text{clip}\big(r_i(\theta), 1-\epsilon, 1+\epsilon\big) A_i \Big)  - \gamma \mathbb{D}_{KL}(\pi_\theta \| \pi_{\text{ref}})\bigg],
\end{split}
\end{equation}
% and KL regularization coefficient, respectively.
where $r_i(\theta) = \frac{\pi_\theta(a_i|x,q)}{\pi_{\text{old}}(a_i|x,q)}$ is the probability ratio, $\pi_{\text{ref}}$ is the frozen reference model, and $\epsilon, \gamma$ are the hyperparameters.
The advantage $A_i$ is computed via group normalization to reduce variance: $A_i = (R(a_i) - \bar{R}) / \sigma_R$, where $\bar{R}$ and $\sigma_R$ denote the mean and standard deviation of rewards within the group. The optimization is guided by a knowledge-aware multi-granular reward function: $R(a_i) = R_{\text{format}}(a_i) + R_{\text{semantic}}(a_i) + \alpha R_{\text{entity}}(a_i)$.

\noindent \textbf{Format Reward} $R_{\text{format}}$\textbf{.} To ensure the model follows the CoT structure, we introduce a binary reward checking for \texttt{<think>}, \texttt{<observe>}, and \texttt{<answer>} tags: 
\begin{equation}
    R_{\text{format}}(a_i) = 
    \begin{cases} 
      1, & \text{if correct format} \\[5pt]
      0, & \text{otherwise}
    \end{cases} .
\end{equation}

\noindent \textbf{Semantic Reward} $R_{\text{semantic}}$\textbf{.} We use GPT-4o as a judge to evaluate the clinical accuracy and logical consistency of the prediction $a_{\text{pred}}$ against the ground truth $a_{\text{gt}}$. The score is continuous: $R_{\text{semantic}}(a_i) = \text{Score}_{\text{LLM}}(a_{\text{pred}}, a_{\text{gt}}) \in [0, 1]$.

% \noindent \textbf{Entity Reward} $R_{\text{entity}}$\textbf{.}
% The KG-based construction of PathReasoner provides explicit entity information, enabling us to formulate a highly effective reward signal. We design $R_{\text{entity}}$ using the Soft-Dice Coefficient to enforce alignment with ground-truth entities.
% Let $E_{\text{pred}}$ and $E_{\text{gt}}$ denote the entity sets from the prediction and ground truth.
% The reward is defined as:
% \begin{equation}
% R_{\text{entity}} = \frac{2 \cdot \mathcal{I}_{\text{soft}}}{|E_{\text{pred}}| + |E_{\text{gt}}| + \epsilon},
% \label{eq:entity}
% \end{equation}
% where $\epsilon$ prevents division by zero, and $\text{sim}(e, e’)$ denotes the cosine similarity between BioBERT embeddings. 
% The $\mathcal{I}_{\text{soft}}$ is a generalized intersection:
% \begin{equation} \mathcal{I}_{\text{soft}} = |E_{\text{pred}} \cap E_{\text{gt}}| + \beta \sum_{e \in E_{\text{pred}} \setminus E_{\text{gt}}} \max_{e’ \in E_{\text{gt}}} \text{sim}(e, e’). \end{equation}
% By weighting unmatched entities with their max-similarity score $\beta$, this metric robustly rewards medically synonymous findings while suppressing non-existent fabrications.

\noindent \textbf{Entity Reward} $R_{\text{entity}}$\textbf{.}
The KG-based construction of PathReasoner provides explicit entity information for a structured reward signal. We define $R_{\text{entity}}$ using a Soft-Dice Coefficient to align predicted and ground-truth entity
sets $E_{\text{pred}}$ and $E_{\text{gt}}$:
\begin{equation}
R_{\text{entity}} = \frac{2 \cdot \mathcal{I}_{\text{soft}}}{|E_{\text{pred}}| + |E_{\text{gt}}| + \epsilon},
\label{eq:entity}
\end{equation}
where $\epsilon$ is a small constant, and the intersection $\mathcal{I}_{\text{soft}}$ is:
\begin{equation} 
\mathcal{I}_{\text{soft}} = |E_{\text{pred}} \cap E_{\text{gt}}| + \beta \sum_{e \in E_{\text{pred}} \setminus E_{\text{gt}}} \max_{e’ \in E_{\text{gt}}} \text{sim}(e, e’).
\end{equation} 
Here, $\text{sim}(\cdot, \cdot)$ computes the cosine similarity of BioBERT embeddings with scaling unmatched entities via $\beta \in [0, 1]$.
The mechanism rewards semantic consistency while suppressing hallucinations or shortcut learning.

\section{Experiments}
We first evaluate PathReasoner-R1 on the PathReasoner testing set (Figure \ref{fig:data_distribution}(a)) to assess its diagnostic performance and CoT generation capabilities. Furthermore, we validate the model's generalization through extensive out-of-domain experiments on external benchmarks across multiple scales, including WSI-level datasets (SlideBench~\cite{SlideChat}, WSI-VQA~\cite{WSI-VQA}, CPTAC~\cite{CPTAC}) and ROI-level datasets (PathMMU~\cite{PathMMU}). Implementation details and evaluation metrics are provided in Appendix~\ref{sec:implementations} and Appendix~\ref{sec:evaluation}.

We compare PathReasoner-R1 with 12 state-of-the-art VLMs, which can be divided into two categories: (1) Non-reasoning models: Quilt-LLaVA \cite{Quilt-LLaVA}, MedGemma-4B-IT \cite{MedGemma}, LLaVA-Med \cite{li2023llava}, Qwen2.5-VL-8B \cite{Qwen2.5}, Qwen3-VL-8B \cite{Qwen3-VL}, HuatuoGPT-Vision \cite{Huatuogpt}, SlideChat \cite{SlideChat} and WSI-LLaVA \cite{WSI-LLaVA}; (2) Reasoning-capable VLMs: Qwen3-VL-8B-Thinking \cite{Qwen3-VL}, InternVL3.5-8B \cite{InternVL}, MedVLThinker-7B \cite{MedVLThinker}, and PathoR1-7B \cite{Patho-R1}.

\subsection{Open-ended Pathology Analysis Evaluation}
Quantitative results on the PathReasoner benchmark are presented in Table~\ref{tab:ours}. Existing VLMs exhibit clear limitations: they are either confined to local patches during reasoning (e.g., Patho-R1) or lack explicit reasoning capabilities (e.g., SlideChat). In contrast, PathReasoner is the only framework capable of performing deep CoT reasoning directly at the slide level. The consistently low scores across baselines highlight the significant challenge of this task. Notably, our supervised model, PathReasoner-SFT-7B, significantly outperforms all existing approaches. This lead validates the value of our dataset in bridging the gap between gigapixel images and clinical logic. Furthermore, PathReasoner-R1-7B establishes a state-of-the-art, boosting the LLM Score to 2.583 and achieving a superior BERT Score of 0.779. These results highlight the synergy between our foundational dataset and the reinforcement strategy, which is crucial for mastering verifiable WSI analysis.

\begin{figure*}[t]
  \centering
  \includegraphics[width=\textwidth]{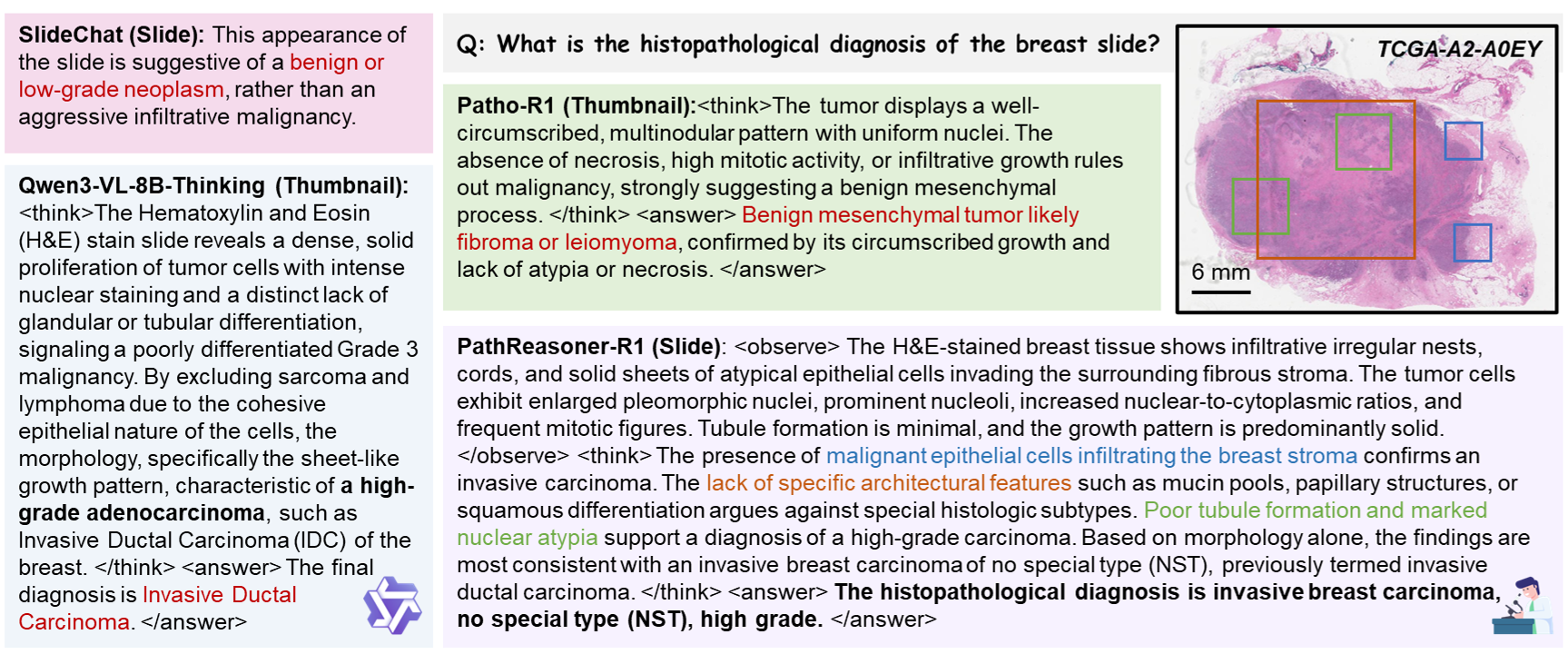}
  \caption{Qualitative comparison of different VLMs for pathology diagnosis. Red text indicates an incorrect diagnosis, while bold text indicates a correct diagnosis. The proposed PathReasoner‑R1 successfully captures pathological visual features (blue, orange, and green texts) and provides a comprehensive reasoning process for accurate slide‑level diagnosis. More samples are in Appendix~\ref{subsec:qualitative}.}
  \label{fig:diagnosis}
\vspace{-10pt}
\end{figure*}

\begin{figure}[t]
    \centering
    \includegraphics[width=\columnwidth]{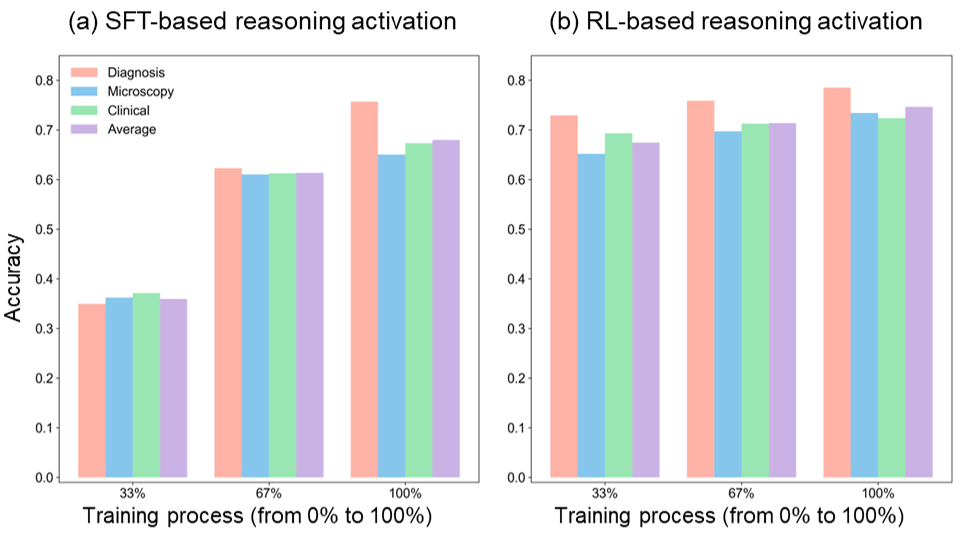}
    \caption{The accuracy changes on the external SlideBench-TCGA benchmark during the two-phase training process.}
    \label{fig:ablation_acc}
\vspace{-15pt}
\end{figure}

\subsection{Generalization Ability Evaluation}
To assess generalization, we evaluate PathReasoner on four external WSI-level benchmarks and one ROI-level benchmark, as detailed in Tables \ref{tab:slide} and \ref{tab:patch}. Appendices~\ref{subsec:patch_level} and \ref{subsec:wsi_level} contain additional results.

\noindent \textbf{WSI-Level Performance.} While the SFT baseline provides a robust foundation, the RL stage in PathReasoner-R1 yields critical improvements, particularly in tasks requiring complex logic. This is most evident on the challenging SlideBench-BCNB: for fine-grained diagnostics such as tumor grading and subtype classification, PathReasoner-R1 outperforms its SFT counterpart by margins of +7.16\% and +8.60\%, respectively. This gap suggests that while visual pattern recognition suffices for simple subtyping, a structured reasoning process is essential for nuanced pathological distinctions.

Consequently, this enhanced reasoning capability translates into state-of-the-art performance. The proposed PathReasoner-R1 achieves a leading average score of 57.68\% on SlideBench-BCNB, surpassing strong baselines like WSI-LLaVA at 55.30\% and significantly exceeding SlideChat at 43.60\%. Although SlideChat retains a marginal advantage on SlideBench-TCGA due to training distribution overlap, our model demonstrates superior robustness on strictly unseen cohorts. Notably, it sets new records on CPTAC and WSI-VQA with accuracies of 74.95\% and 55.90\%, respectively, validating that our model acquires transferable clinical logic rather than relying on rote memorization.

\noindent \textbf{ROI-Level Performance.} Beyond slide-level analysis, we validated PathReasoner-R1 on PathMMU to assess its region understanding ability. As shown in Table~\ref{tab:patch}, PathReasoner-R1 demonstrates competitive performance in identifying local pathological features. This indicates that the reasoning capabilities developed at the WSI level effectively transfer to ROI-level tasks, enabling the model to maintain high diagnostic precision, establishing PathReasoner as a versatile solution for CPath.

\subsection{Reasoning Capability Evaluation}
To evaluate reasoning capability, we compare the quality and coherence of our method's chain-of-thought generation with state-of-the-art baselines using GPT-4o-based evaluation. In Table \ref{tab:ours}, PathReasoner-R1 outperforms Patho-R1, achieving an 8.1\% higher A-score in alignment accuracy and a 5.4\% improvement in Q-score for reasoning step quality.

This advantage is particularly pronounced in ambiguous cases where baselines suffer from severe hallucinations. As visualized in Figure~\ref{fig:diagnosis}, Patho-R1 and SlideChat failed to identify the malignancy, erroneously classifying the tissue as a benign neoplasm due to hallucinations of ``uniform nuclei.'' Meanwhile, although Qwen3-VL-8B-Thinking suspects malignancy, it produces false histological evidence (e.g., glandular differentiation). PathReasoner-R1 avoids these pitfalls by strictly grounding its reasoning in observed visual features. By logically excluding specific subtypes based on the absence of architectural patterns (e.g., mucin pools), our method achieves a diagnosis consistent with the ground truth, validating that high-quality CoT is essential for mitigating visual hallucinations.

\subsection{Training Dynamics Performance}
To monitor the model's performance during training, we evaluated the model on the SlideBench-TCGA benchmark at multiple checkpoints for both training stages. As illustrated in Figure~\ref{fig:ablation_acc}, the SFT phase exhibits a sharp upward trajectory, with accuracy nearly doubling from 33\% to 100\% of the training progress. This indicates that the model rapidly acquired domain-specific instruction-following capabilities. Subsequently, the RL stage builds upon this foundation. Starting with a substantial baseline, the model demonstrates steady, continuous gains across all subtasks, particularly in diagnosis. This validates that our knowledge-guided RL effectively refines the model's reasoning logic.

\begin{table}[t]
\caption{Ablation study of RL hyperparameters $\alpha$ and $\beta$ on SlideBench-TCGA across different subsets. The first row ``$-$'' represents the SFT baseline. Best results are in \textbf{bold}.}
\centering
\renewcommand{\arraystretch}{1.20}
\setlength{\tabcolsep}{4pt}
\small
\begin{tabular}{cccccc}
\toprule
\toprule
\multicolumn{2}{c}{Hyperparameters} & \multicolumn{4}{c}{SlideBench-TCGA} \\
\cmidrule(lr){1-2} \cmidrule(lr){3-6}
$\alpha$ & $\beta$ & Microscopy & Diagnosis & Clinical & Average \\
\midrule
- & - & 75.73 & 65.05 & 67.35 & 67.98 \\
0 & 0 & 73.23 & 67.64 & 69.13 & 69.19 \\
1.0 & 0 & 74.25 & 70.80 & \textbf{74.95} & 71.97 \\
1.0 & 0.5 & \textbf{78.52} & \textbf{73.43} & 72.39 & \textbf{74.68} \\
\bottomrule
\bottomrule
\end{tabular}
\label{table_ablation}
\vspace{-18pt}
\end{table}

\subsection{Ablation on Entity Reward} Table~\ref{table_ablation} presents the ablation study on the entity reward mechanism. Compared to the RL baseline without entity supervision, integrating entity rewards with $\alpha$ set to 1.0 and $\beta$ to 0.5 yields a 5.49\% improvement in average accuracy, reaching an overall accuracy of 74.68\%. This significant gain confirms that explicitly aligning reasoning paths with medical entities is crucial for accurate diagnosis. Notably, the Microscopy score rises from 74.25\% to 78.52\%, suggesting that the soft-matching coefficient $\beta=0.5$ effectively captures fine-grained visual features by accommodating synonymous variations in pathological descriptions.

\section{Conclusion}
We introduce PathReasoner-R1 to bridge the critical gap between visual perception and clinical logic in CPath. By leveraging the constructed large-scale WSI reasoning dataset and a knowledge-guided learning paradigm, our method equips VLMs with structured, verifiable CoT capabilities. PathReasoner-R1 not only achieves state-of-the-art performance but also ensures diagnostic transparency through evidence-based rationales. This work signifies a pivotal shift from imitation-based learning to genuine reasoning, establishing a new foundation for trustworthy CPath.

\section{Impact Statement}
This work aims to improve the transparency and trustworthiness of computational pathology systems. By grounding model outputs in verifiable medical knowledge graphs and structured reasoning, we strive to reduce the risk of unfounded hallucinations common in generative models. While this contributes positively to clinical decision support, we acknowledge that any AI system deployed in healthcare must undergo strict clinical validation and operate under human supervision to prevent potential misuse or over-reliance on automated diagnoses.

\nocite{langley00}
\bibliography{reference}
\bibliographystyle{icml2026}

%%%%%%%%%%%%%%%%%%%%%%%%%%%%%%%%%%%%%%%%%%%%%%%%%%%%%%%%%%%%%%%%%%%%%%%%%%%%%%%
%%%%%%%%%%%%%%%%%%%%%%%%%%%%%%%%%%%%%%%%%%%%%%%%%%%%%%%%%%%%%%%%%%%%%%%%%%%%%%%
% APPENDIX
%%%%%%%%%%%%%%%%%%%%%%%%%%%%%%%%%%%%%%%%%%%%%%%%%%%%%%%%%%%%%%%%%%%%%%%%%%%%%%%
%%%%%%%%%%%%%%%%%%%%%%%%%%%%%%%%%%%%%%%%%%%%%%%%%%%%%%%%%%%%%%%%%%%%%%%%%%%%%%%
\newpage
\appendix
\onecolumn
\section*{Appendix}
\textbf{Table of content:}
\begin{itemize}
\item \S\ref{sec:construction}: Details of PathReasoner Construction
    \begin{itemize}
    \item \S\ref{subsec:kg}: Knowledge Graph Construction
    \item \S\ref{subsec:entity}: Reasoning Construction Pipeline
    \item \S\ref{subsec:trajectory}: Mask Trajectory Sampling
    \end{itemize}
    
\item \S\ref{sec:implementations}: Implementation details of PathReasoner-R1
    \begin{itemize}
    \item \S\ref{subsec:baseline}: Baseline of PathReasoner-R1
    \item \S\ref{subsec:preprocess}: WSI Pre-Process in PathReasoner-R1
    \item \S\ref{subsec:training}: Training Configurations
    \end{itemize}
\item \S\ref{sec:evaluation}: Evaluation Metrics
\item \S\ref{sec:comparison}: More comparison experiments and ablation studies
    \begin{itemize}
    \item \S\ref{subsec:prompts}: Prompts used to generation and evaluation
    \item \S\ref{subsec:patch_level}: Evaluation on Patch-Level Benchmarks
    \item \S\ref{subsec:wsi_level}: Evaluation on WSI-Level Benchmarks
    \item \S\ref{subsec:rewards}: Training Dynamics Rewards Performance
    \item \S\ref{subsec:sampling}: Impact of Trajectory Augmentation
    \item \S\ref{subsec:qualitative}: Qualitative Results
    \end{itemize}

\end{itemize}
\section{Details of PathReasoner Construction}
\label{sec:construction}

\subsection{Knowledge Graph Construction}
\label{subsec:kg}

\textbf{Data Sources and Integration.} To establish a comprehensive foundation for pathology reasoning, we constructed a multi-scale pathology knowledge graph $\mathcal{G}$ by integrating two authoritative sources:
(i) PrimeKG (Macro-scale Context)~\cite{PrimeKG}: Providing high-level medical context, this graph contains over 4 million edges linking diseases to molecular signatures (e.g., genes, proteins) and clinical phenotypes. We leveraged PrimeKG’s structured disease-phenotype relationships to ground diagnostic targets within established precision medicine knowledge.
(ii) PathoGraph (Micro-scale Topology)~\cite{pathograph}: Offering specialized histopathological structure, PathoGraph represents tissue sections as a hierarchical graph where nodes correspond to physical entities (e.g., cell nuclei, stroma) and edges encode spatial proximity, morphological attributes, and diagnostic evidence. It captures the transition from low-level visual features to high-level pathology findings.

\textbf{Graph Alignment and Fusion.} The construction process bridges the gap between PathoGraph’s microscopic clusters and PrimeKG’s clinical disease nodes:
(i) Node Alignment: We mapped \textit{Diagnosis} and \textit{Disease} nodes from PathoGraph to corresponding \textit{Disease} nodes in PrimeKG to ensure a unified semantic space. Specifically, 85\% of nodes were aligned via exact matching of UMLS/MONDO IDs, while the remaining terms were aligned using cosine similarity of their BioBERT embeddings (threshold $> 0.85$).
(ii) Relational Linking: We expanded the graph by establishing edges between PathoGraph’s \textit{Pathology Phenotype} nodes and PrimeKG’s \textit{Clinical Phenotype} nodes. This bridge creates a structural continuity from micro-scale morphological changes (e.g., nuclear atypia) to systemic clinical symptoms.
(iii) Topology Integration for Reasoning: The fusion creates a unified topology optimized for algorithmic path retrieval. This structural integration ensures the existence of computable visual-to-clinical pathways that span from micro-scale spatial entities in PathoGraph, through interpretative logic layers, to macro-scale biological implications in PrimeKG. To ensure robust global connectivity for pathfinding, redundant edges were pruned, and isolated subgraphs were removed. The specific composition of nodes and edges across different scales is listed in Table~\ref{tab:kg_stats}.

\begin{table}[htbp]
\centering
\caption{Summary statistics of the constructed Pathology Knowledge Graph $\mathcal{G}$.}
\label{tab:kg_stats}
\renewcommand{\arraystretch}{1.2} 
\setlength{\tabcolsep}{10pt} 
\begin{tabular}{ccc} 
\toprule
\textbf{Scale \& Category} & \textbf{Counts} & \textbf{Examples} \\
\midrule
\multicolumn{3}{l}{\textit{\textbf{Micro-scale (Pathological Concepts)}}} \\
\quad Histological Entities & 120 & Nuclei, Stroma, Glands \\
\quad Visual Phenotypes & 85 & Atypia, Mitosis, Necrosis \\
\midrule
\multicolumn{3}{l}{\textit{\textbf{Macro-scale (Clinical Context)}}} \\
\quad Diseases & 1,500 & Lung Adenocarcinoma, SCC \\
\quad Genes/Proteins & 3,200 & EGFR, KRAS, TP53 \\
\quad Clinical Phenotypes & 2,500 & Cough, Dyspnea \\
\midrule 
\multicolumn{3}{l}{\textit{\textbf{Graph Topology}}} \\ 
\quad \textbf{Total Nodes} & \textbf{7,405} & \textit{(Sum of unique entities)} \\
\quad \textbf{Total Edges} & \textbf{45,200} & --- \\
\quad \textbf{Relation Types} & \textbf{25} & \textit{indicated\_by, associated\_with} \\ 
\bottomrule
\end{tabular}
\end{table}
%%

% 展示我们用的一些采样、构造、提示词，
% 展示我们轨迹mask前后的文本差异。
% 展示一些数据集的对比

\subsection{Reasoning Construction Pipeline}
\label{subsec:entity}
The pipeline, illustrated in Figures \ref{ReasonPipeline1}--\ref{ReasonPipeline_last}, leverages our Knowledge Graph $\mathcal{G}$ to generate verifiable diagnostic chains through a three-step process. 
(i) Entity Anchoring: We utilize GPT-4o to extract context-aware entities from diagnostic reports, mapping them to specific nodes in $\mathcal{G}$ (e.g., \textit{Physical\_Entity}, \textit{Phenotype}).
(ii) Path Retrieval: Using these entities as anchors, we employ a shortest-path retrieval algorithm to identify the most concise logical trajectories between nodes. This process prioritizes edges representing core diagnostic logic (e.g., \textit{hasSupportEvidence}), capturing both direct causal links and the nuances of differential diagnosis.
(iii) Grounded Generation: The retrieved graph trajectories are injected into the LLM context. The model then synthesizes these structured paths into a natural language Chain of Thought. This process ensures that the generated reasoning is strictly bound by the medical facts in $\mathcal{G}$, enabling the creation of a high-fidelity dataset for distilling reasoning capabilities into smaller models such as with 3B--8B parameters.

%% CoT Generation
\begin{figure}[!p]
    \centering
    \begin{tikzpicture}
        % 主内容框
        \node[
            draw=framecolor,
            line width=1.5pt,
            rounded corners=8pt,
            inner sep=12pt,
            text width=16cm, % 稍微加宽一点
            align=left,
            font=\small\sffamily,
            fill=codebg
        ] (box) {
            
            % ==========================================
            % PHASE 1: EXTRACTION
            % ==========================================
            {
            \ttfamily
            \textbf{\textcolor{keywordcolor}{SYSTEM}:} You are a Pathological Entity Extraction Engine. Extract terms strictly adhering to the schema: [\texttt{Physical\_Entity}, \texttt{Phenotype}, \texttt{Diagnosis}]. \\
            \vspace{0.2em}
            \textbf{\textcolor{keywordcolor}{USER}:} Analyze the following pathology report. Identify precise mention spans, map them to the schema types, and output the result in JSON format.
            }
            
            \vspace{0.3em}
            \textbf{--- Input Text (Excerpt) ---} \\
            \textit{"...The primary tumor is located in the \underline{main bronchus}. ... \underline{Tumor cells} \underline{invading} through the \underline{basement membrane} is a key feature of \underline{invasive squamous cell carcinoma}..."}
            
            \vspace{0.3em}
            \textbf{--- Step 1: Model Output  ---} \\

            \ttfamily 
            \{\{ \\
            \hspace*{1em} "extracted\_entities": [ \\
            \hspace*{2em} \textcolor{gray}{// 1. Physical Entities} \\
            \hspace*{2em} \{\{"id": "E1", "name": "\textbf{Main bronchus}", "type": "Structure"\}\}, \\
            \hspace*{2em} \{\{"id": "E2", "name": "\textbf{Basement membrane}", "type": "Structure"\}\}, \\
            \hspace*{2em} \{\{"id": "E3", "name": "\textbf{Tumor cells}", "type": "Structure"\}\}, \\
            
            \vspace{0.2em}
            \hspace*{2em} \textcolor{gray}{// 2. Pathological Phenotypes} \\
            \hspace*{2em} \{\{"id": "P1", "name": "\textbf{Invading}", "type": "Phenotype"\}\}, \\
            \hspace*{2em} \{\{"id": "P2", "name": "\textbf{Metastasis}", "type": "Phenotype"\}\}, \\
            \hspace*{2em} \{\{"id": "P3", "name": "\textbf{Glandular structures}", "type": "Phenotype"\}\}, \\
            
            \vspace{0.2em}
            \hspace*{2em} \textcolor{gray}{// 3. Diagnostic Concepts} \\
            \hspace*{2em} \{\{"id": "D1", "name": "\textbf{Squamous cell carcinoma}", "type": "Diagnosis"\}\}, \\
            \hspace*{2em} \{\{"id": "D2", "name": "\textbf{Adenocarcinoma}", "type": "Diagnosis"\}\} \\
            \hspace*{1em} ] \\
            \}\}
            \vspace{0.5em}
            % ==========================================
            % SEPARATOR 1: Extraction -> Retrieval
            % ==========================================
            % \hrule height 0.5pt \vspace{0.1em}
            \begin{center}
                \scriptsize\color{gray} $\Downarrow$ Querying Graph with Extracted Entities $\Downarrow$
            \end{center}
            \vspace{1em}
            % ==========================================
            % PHASE 2: RETRIEVAL
            % ==========================================
            \normalfont\sffamily
            \textbf{--- Step 2: Retrieved Knowledge Paths ---} \\
            \vspace{0.2em}
            \ttfamily\footnotesize
            
            \textbf{Path A (Anatomical Context):} \\
            \hspace*{1em} [Main Bronchus] \textcolor{gray}{--hasComponent$\to$} [Epithelium] \textcolor{gray}{--adjacentTo$\to$} [Basement Membrane] \\
            
            \vspace{0.2em}
            \textbf{Path B (Diagnostic Evidence):} \\
            \hspace*{1em} [Basement Membrane] \textcolor{gray}{--siteOf$\to$} [Invasion] \textcolor{gray}{--keyFeatureOf$\to$}[Squamous Cell Carcinoma] \\
            
            \vspace{0.2em}
            \textbf{Path C (Differential Pattern):} \\
            \hspace*{1em} [Adenocarcinoma]\textcolor{gray}{--manifests$\to$} [Glandular Structures] \\

            \vspace{0.2em}
            \textbf{Path D (Exclusion Criteria):} \\
            \hspace*{1em} [Squamous Cell Carcinoma] \textcolor{gray}{--excludes$\to$} [Glandular Structures]
            
            \vspace{1em}
            % ==========================================
            % SEPARATOR 2: Retrieval -> Reasoning
            % ==========================================
            % \hrule height 0.5pt \vspace{0.1em}
            \begin{center}
                \scriptsize\color{gray} $\Downarrow$ Injecting Paths into Reasoning Engine $\Downarrow$
            \end{center}
            \vspace{1em}

            % ==========================================
            % PHASE 3: REASONING
            % ==========================================
            \normalfont\sffamily
            \textbf{--- Step 3: LLM-Integrated Chain of Thought ---} \\
            {
            \ttfamily
            \textbf{\textcolor{keywordcolor}{SYSTEM}:} You are a Knowledge-Grounded Reasoning Assistant. Your task is to synthesize a diagnostic rationale based on the provided Knowledge Graph paths. \\
            \vspace{0.2em}
            \textbf{\textcolor{keywordcolor}{USER}:}  Synthesize these paths into a coherent diagnostic chain. Explicitly state how the morphological evidence supports the Final Diagnosis and excludes alternatives.
            }
            \vspace{0.4em}
            \vspace{0.2em}

            \textbf{Reasoning Output:} \\
            {
            \linespread{1.1}\selectfont 
            \small \itshape 
            ``Histologically, the lesion is localized within the \underline{Main Bronchus}, where the architectural relationship between the surface epithelium and the \underline{adjacent} \underline{Basement Membrane} is scrutinized. 
            
            A defining pathological event is observed here: the \underline{Basement Membrane} serves as the direct \underline{site of} Invasion by neoplastic cells. 
            This breach is clinically significant because such invasion is a \underline{key feature} characteristic of \underline{Squamous Cell Carcinoma}. 
            
            For differential diagnosis, we distinguish this phenotype from \underline{Adenocarcinoma}. 
            The logic relies on morphological patterns: whereas \underline{Adenocarcinoma} predictably \underline{manifests} \underline{Glandular Structures}, the definition of \underline{Squamous Cell Carcinoma} explicitly \underline{excludes} them. 
            
            Therefore, the final diagnosis answer is squamous cell carcinoma."
            \par 
            }
            \vspace{0.4em}
        };
        
        % 标题 Tab
        \node[
            fill=framecolor,
            text=white,
            rounded corners={5pt},
            anchor=west,
            xshift=20pt
        ] at (box.north west) {
            \bfseries \sffamily \hspace{0.5em} Reasoning Generation Pipeline \hspace{0.5em}
        };
        
    \end{tikzpicture}
    
    \caption{\textbf{Full Reasoning Pipeline (Squamous Cell Carcinoma Pathology).} Step 1 begins with schema-guided entity extraction. The extracted entities act as anchors to retrieve relevant structural and diagnostic paths from the Knowledge Graph in Step 2. Finally,  Step 3 provides the grounded context for the LLM to generate a verified Chain of Thought.}
\label{ReasonPipeline1}
\end{figure}

\begin{figure}[!p]
    \centering
    \begin{tikzpicture}
        % 主内容框
        \node[
            draw=framecolor,
            line width=1.5pt,
            rounded corners=8pt,
            inner sep=12pt,
            text width=16cm, 
            align=left,
            font=\small\sffamily,
            fill=codebg
        ] (box) {
            
            % ==========================================
            % PHASE 1: EXTRACTION
            % ==========================================
            {
            \ttfamily
            \textbf{\textcolor{keywordcolor}{SYSTEM}:} You are a Pathological Entity Extraction Engine. Extract terms strictly adhering to the schema: [\texttt{Physical\_Entity}, \texttt{Phenotype}, \texttt{Diagnosis}]. \\
            \vspace{0.2em}
            \textbf{\textcolor{keywordcolor}{USER}:} Analyze the following pathology report. Identify precise mention spans, map them to the schema types, and output the result in JSON format.
            }
            
            \vspace{0.3em}
            \textbf{--- Input Text (Excerpt) ---} \\
            \textit{"...The tumor demonstrates \underline{high nuclear grade} \underline{ductal carcinoma in-situ} ... The surrounding \underline{invasive component} displays \underline{marked nuclear pleomorphism}, numerous mitotic figures, and a \underline{lack of tubule formation}, consistent with \underline{Nottingham Grade III}..."}
            
            \vspace{0.3em}
            \textbf{--- Step 1: Model Output  ---} \\

            \ttfamily 
            \{\{ \\
            \hspace*{1em} "extracted\_entities": [ \\
            \hspace*{2em} \textcolor{gray}{// 1. Physical Entities} \\
            \hspace*{2em} \{\{"id": "E1", "name": "\textbf{Ductal carcinoma in-situ}", "type": "Structure"\}\}, \\
            \hspace*{2em} \{\{"id": "E2", "name": "\textbf{Invasive component}", "type": "Structure"\}\}, \\
            
            \vspace{0.2em}
            \hspace*{2em} \textcolor{gray}{// 2. Pathological Phenotypes} \\
            \hspace*{2em} \{\{"id": "P1", "name": "\textbf{Marked nuclear pleomorphism}", "type": "Phenotype"\}\}, \\
            \hspace*{2em} \{\{"id": "P2", "name": "\textbf{Lack of tubule formation}", "type": "Phenotype"\}\}, \\
            \hspace*{2em} \{\{"id": "P3", "name": "\textbf{Stromal invasion}", "type": "Phenotype"\}\}, \\
            
            \vspace{0.2em}
            \hspace*{2em} \textcolor{gray}{// 3. Diagnostic Concepts} \\
            \hspace*{2em} \{\{"id": "D1", "name": "\textbf{Invasive Ductal Carcinoma}", "type": "Diagnosis"\}\}, \\
            \hspace*{2em} \{\{"id": "D2", "name": "\textbf{Nottingham Grade III}", "type": "Diagnosis"\}\} \\
            \hspace*{1em} ] \\
            \}\}
            \vspace{0.5em}
            % ==========================================
            % SEPARATOR 1: Extraction -> Retrieval
            % ==========================================
            % \hrule height 0.5pt \vspace{0.1em}
            \begin{center}
                \scriptsize\color{gray} $\Downarrow$ Querying Graph with Extracted Entities $\Downarrow$
            \end{center}
            \vspace{1em}
            % ==========================================
            % PHASE 2: RETRIEVAL
            % ==========================================
            \normalfont\sffamily
            \textbf{--- Step 2: Retrieved Knowledge Paths ---} \\
            \vspace{0.2em}
            \ttfamily\footnotesize
            
            \textbf{Path A (Diagnosis):} \\
            \hspace*{1em} [Invasive Component] \textcolor{gray}{--invades$\to$} [Stroma] \textcolor{gray}{--distinguishesFrom$\to$} [Ductal carcinoma in-situ] \\
            
            \vspace{0.2em}
            \textbf{Path B (Grading Criteria):} \\
            \hspace*{1em} [Marked nuclear pleomorphism] \textcolor{gray}{--scoredAs$\to$} [Score 3 (Nuclear)] \\
            \hspace*{1em} [Lack of tubule formation] \textcolor{gray}{--scoredAs$\to$} [Score 3 (Tubule)] \\
            \hspace*{1em} [High mitotic rate] \textcolor{gray}{--scoredAs$\to$} [Score 3 (Mitosis)] \\
            
            \vspace{0.2em}
            \textbf{Path C (Conclusion):} \\
            \hspace*{1em} [Score 3 (Nuclear)] \textcolor{gray}{--plus$\to$} [Score 3 (Tubule)]  \textcolor{gray}{--plus$\to$} [Score 3 (Mitosis)] \textcolor{gray}{--resultsIn$\to$} [High Total Score] \textcolor{gray}{--classifiesAs$\to$} [Nottingham Grade III]

            \vspace{1em}
            % ==========================================
            % SEPARATOR 2: Retrieval -> Reasoning
            % ==========================================
            % \hrule height 0.5pt \vspace{0.1em}
            \begin{center}
                \scriptsize\color{gray} $\Downarrow$ Injecting Paths into Reasoning Engine $\Downarrow$
            \end{center}
            \vspace{1em}

            % ==========================================
            % PHASE 3: REASONING
            % ==========================================
            \normalfont\sffamily
            \textbf{--- Step 3: LLM-Integrated Chain of Thought ---} \\
            {
            \ttfamily
            \textbf{\textcolor{keywordcolor}{SYSTEM}:} You are a Knowledge-Grounded Reasoning Assistant. Your task is to synthesize a diagnostic rationale based on the provided Knowledge Graph paths. \\
            \vspace{0.2em}
            \textbf{\textcolor{keywordcolor}{USER}:}  Synthesize these paths into a coherent diagnostic chain. Explicitly state how the morphological evidence supports the Final Diagnosis and excludes alternatives.
            }
            \vspace{0.4em}
            \vspace{0.2em}

            \textbf{Reasoning Output:} \\
            {
            \linespread{1.1}\selectfont 
            \small \itshape 
            ``Morphologically, the presence of \underline{Ductal carcinoma in-situ} suggests the tumor origin. However, the critical feature is the \underline{Invasive component} extending into the stroma, which \underline{invades} the surrounding tissue and \underline{distinguishes from} pure in-situ disease, confirming \textbf{Invasive Ductal Carcinoma}. 
            
            To determine the grade, we evaluate specific H\&E features: the \underline{Marked nuclear pleomorphism} indicates high-grade atypia and is \underline{scored as} Score 3. Similarly, the \underline{Lack of tubule formation} represents a solid growth pattern, which is also \underline{scored as} Score 3. 
            The presence of High mitotic activity  \underline{contributes} a Score of 3.
            
            Aggregating these morphological scores \underline{results in} a high total score (8-9), which strictly \underline{classifies as} \textbf{Nottingham Grade III}. Thus, the final diagnosis is high-grade invasive ductal carcinoma."
            \par 
            }
            \vspace{0.4em}
        };
        
        % 标题 Tab
        \node[
            fill=framecolor,
            text=white,
            rounded corners={5pt},
            anchor=west,
            xshift=20pt
        ] at (box.north west) {
            \bfseries \sffamily \hspace{0.5em} Reasoning Generation Pipeline \hspace{0.5em}
        };
        
    \end{tikzpicture}
    
    \caption{\textbf{Full Reasoning Pipeline (Breast Pathology).} Step 1 begins with extracting morphological entities from the H\&E description. Step 2 retrieves grading criteria paths from the Knowledge Graph , connecting pleomorphism and tubule formation to Nottingham scores. Finally, the LLM synthesizes these paths into a cohesive diagnosis of Invasive Ductal Carcinoma, Grade III in Step 3.}
\label{ReasonPipeline2}
\end{figure}

\begin{figure}[!p]
    \centering
    \resizebox{1.0\textwidth}{!}{  % 强制宽度为文本宽，高度按比例缩放
    \begin{tikzpicture}
        % 主内容框
        \node[
            draw=framecolor,
            line width=1.5pt,
            rounded corners=8pt,
            inner sep=12pt,
            text width=16cm, 
            align=left,
            font=\small\sffamily,
            fill=codebg
        ] (box) {
            
            % ==========================================
            % PHASE 1: EXTRACTION
            % ==========================================
            {
            \ttfamily
            \textbf{\textcolor{keywordcolor}{SYSTEM}:} You are a Pathological Entity Extraction Engine. Extract terms strictly adhering to the schema: [\texttt{Physical\_Entity}, \texttt{Phenotype}, \texttt{Diagnosis}]. \\
            \vspace{0.2em}
            \textbf{\textcolor{keywordcolor}{USER}:} Analyze the following pathology report. Identify precise mention spans, map them to the schema types, and output the result in JSON format.
            }
            
            \vspace{0.3em}
            \textbf{--- Input Text (Excerpt) ---} \\
            \textit{"...solid sheets of \underline{pleomorphic cells} with \underline{numerous mitotic figures} ... ducts expanded by \underline{central expansive necrosis} ... clusters of tumor cells within \underline{endothelium-lined vascular spaces}..."}
            
            \vspace{0.3em}
            \textbf{--- Step 1: Model Output (Expanded Entities) ---} \\

            \ttfamily 
            \{\{ \\
            \hspace*{1em} "extracted\_entities": [ \\
            \hspace*{2em} \textcolor{gray}{// 1. Physical Entities} \\
            \hspace*{2em} \{\{"id": "E1", "name": "\textbf{Solid sheets}", "type": "Structure"\}\}, \\
            \hspace*{2em} \{\{"id": "E2", "name": "\textbf{Central necrosis}", "type": "Structure"\}\}, \\
            \hspace*{2em} \{\{"id": "E3", "name": "\textbf{Vascular spaces}", "type": "Structure"\}\}, \\
            \hspace*{2em} \{\{"id": "E4", "name": "\textbf{Endothelium}", "type": "Structure"\}\}, \\
            \hspace*{2em} \{\{"id": "E5", "name": "\textbf{Numerous mitotic figures}", "type": "Structure"\}\}, \\
            
            \vspace{0.2em}
            \hspace*{2em} \textcolor{gray}{// 2. Pathological Phenotypes} \\
            \hspace*{2em} \{\{"id": "P1", "name": "\textbf{Pleomorphism}", "type": "Phenotype"\}\}, \\
            \hspace*{2em} \{\{"id": "P2", "name": "\textbf{Lack of tubules}", "type": "Phenotype"\}\}, \\
            \hspace*{2em} \{\{"id": "P3", "name": "\textbf{Comedo pattern}", "type": "Phenotype"\}\}, \\
            
            \vspace{0.2em}
            \hspace*{2em} \textcolor{gray}{// 3. Diagnostic Concepts} \\
            \hspace*{2em} \{\{"id": "D1", "name": "\textbf{Invasive Ductal Carcinoma}", "type": "Diagnosis"\}\}, \\
            \hspace*{2em} \{\{"id": "D2", "name": "\textbf{Lymphovascular Invasion (LVI)}", "type": "Diagnosis"\}\} \\
            \hspace*{1em} ] \\
            \}\}
            % \vspace{0.5em}
            % ==========================================
            % SEPARATOR 1: Extraction -> Retrieval
            % ==========================================
            \begin{center}
                \scriptsize\color{gray} $\Downarrow$ Querying Graph with Extended Entity Set $\Downarrow$
            \end{center}
            % \vspace{1em}
            % \vspace{0.1em}
            % ==========================================
            % PHASE 2: RETRIEVAL
            % ==========================================
            \normalfont\sffamily
            \textbf{--- Step 2: Retrieved Knowledge Paths (Multi-branch) ---} \\
            % \vspace{0.1em}
            \ttfamily\footnotesize
            
            \textbf{Path A (Tumor Type):} \\
            \hspace*{1em} [Solid sheets] \textcolor{gray}{--lackOf$\to$} [Tubules] \textcolor{gray}{--characteristicOf$\to$} [Invasive Ductal Carcinoma] \\
            
            \textbf{Path B (Grading):} \\
            \hspace*{1em} [Pleomorphism] \textcolor{gray}{--indicates$\to$} [Nuclear Grade 3] \textcolor{gray}{--contributesTo$\to$} [Nottingham Grade III] \\
            \hspace*{1em} [Numerous mitoses]  \textcolor{gray}{--scoredAs$\to$} [Mitotic Score 3] \textcolor{gray}{--contributesTo$\to$} [Nottingham Grade III] \\
            
            \textbf{Path C (In-Situ Component):} \\
            \hspace*{1em} [Central necrosis] \textcolor{gray}{--defines$\to$} [Comedo pattern] \textcolor{gray}{--subtypeOf$\to$} [High Grade DCIS] \\

            \textbf{Path D (Prognostic Feature):} \\
            \hspace*{1em} [Tumor cells] \textcolor{gray}{--inside$\to$} [Vascular spaces] \textcolor{gray}{--linedBy$\to$} [Endothelium] \textcolor{gray}{--confirms$\to$} [LVI]

            % \vspace{0.1em}
            % ==========================================
            % SEPARATOR 2: Retrieval -> Reasoning
            % ==========================================
            \begin{center}
                \scriptsize\color{gray} $\Downarrow$ Injecting Paths into Reasoning Engine $\Downarrow$
            \end{center}
            % \vspace{0.1em}

            % ==========================================
            % PHASE 3: REASONING
            % ==========================================
            \normalfont\sffamily
            \textbf{--- Step 3: LLM-Integrated Chain of Thought ---} \\
            {
            \ttfamily
            \textbf{\textcolor{keywordcolor}{SYSTEM}:} You are a Knowledge-Grounded Reasoning Assistant. Your task is to synthesize a diagnostic rationale based on the provided Knowledge Graph paths. \\
            \vspace{0.2em}
            \textbf{\textcolor{keywordcolor}{USER}:}  Synthesize these paths into a coherent diagnostic chain. Explicitly state how the morphological evidence supports the Final Diagnosis and excludes alternatives.
            }
            % \vspace{0.4em}
            \vspace{0.2em}

            \textbf{Reasoning Output:} \\
            {
            \linespread{1.1}\selectfont 
            \small \itshape 
            ``The histological architecture presents primarily as \underline{Solid sheets} with a conspicuous \underline{Lack of tubules}, which structurally characterizes the lesion as \textbf{Invasive Ductal Carcinoma}. 
            
            Simultaneously, the nuclear features show marked \underline{Pleomorphism} and \underline{Numerous mitoses}, establishing \underline{Nuclear Grade 3} and \underline{Mitotic Grade 3}, which drives the final classification to \textbf{Nottingham Grade III}.
            
            Two critical associated features are identified via specific pathways:
            First, the presence of \underline{Central necrosis} within the ducts strictly \underline{defines} the \textbf{Comedo pattern}, indicating an associated High-Grade DCIS component.
            Second, and most critically for prognosis, tumor clusters are located \underline{inside} \underline{Vascular spaces} that are \underline{lined by} \underline{Endothelium}. This specific anatomical relationship \underline{confirms} the diagnosis of \textbf{Lymphovascular Invasion (LVI)}.
            
            Final Conclusion: High-grade Invasive Ductal Carcinoma with Comedo-DCIS and LVI."
            \par 
            }
            \vspace{0.4em}
        };
        
        % 标题 Tab
        \node[
            fill=framecolor,
            text=white,
            rounded corners={5pt},
            anchor=west,
            xshift=20pt
        ] at (box.north west) {
            \bfseries \sffamily \hspace{0.5em} Reasoning Generation Pipeline \hspace{0.5em}
        };
        
    \end{tikzpicture}
    }
    \caption{\textbf{Complex Reasoning Pipeline (IDC with LVI).} This updated pipeline handles multiple diagnostic tracks simultaneously. Step 1 extracts diverse entities including stromal and vascular components. Step 2 retrieves four distinct logic paths (Type, Grade, Subtype, Invasion). Step 3 synthesizes these into a coherent diagnosis that identifies not just the cancer type, but its specific aggressive features (Comedo necrosis and Lymphovascular Invasion).}
\label{ReasonPipeline3}
\end{figure}

% TCGA-E9-A1RI-01Z-00-DX1.3259644B-0888-4B18-B16F-397D55167EAD
\begin{figure}[!p]
    \centering
    \begin{tikzpicture}
        % 主内容框
        \node[
            draw=framecolor,
            line width=1.5pt,
            rounded corners=8pt,
            inner sep=12pt,
            text width=16cm, 
            align=left,
            font=\small\sffamily,
            fill=codebg
        ] (box) {
            
            % ==========================================
            % PHASE 1: EXTRACTION
            % ==========================================
            {
            \ttfamily
            \textbf{\textcolor{keywordcolor}{SYSTEM}:} You are a Pathological Entity Extraction Engine. Extract terms strictly adhering to the schema: [\texttt{Physical\_Entity}, \texttt{Phenotype}, \texttt{Diagnosis}]. \\
            \vspace{0.2em}
            \textbf{\textcolor{keywordcolor}{USER}:} Analyze the following pathology report. Identify precise mention spans, map them to the schema types, and output the result in JSON format.
            }
            
            \vspace{0.3em}
            \textbf{--- Input Text (Excerpt) ---} \\
            \textit{"...solid sheets of \underline{atypical epithelial cells} with brisk mitotic activity ... dense stromal \underline{inflammatory infiltrate} consisting of \underline{lymphocytes} ... foci of \underline{central comedo-necrosis}..."}
            
            \vspace{0.3em}
            \textbf{--- Step 1: Model Output (Rich Entity Set) ---} \\

            \ttfamily 
            \{\{ \\
            \hspace*{1em} "extracted\_entities": [ \\
            \hspace*{2em} \textcolor{gray}{// 1. Physical Entities} \\
            \hspace*{2em} \{\{"id": "E1", "name": "\textbf{Epithelial tumor cells}", "type": "Structure"\}\}, \\
            \hspace*{2em} \{\{"id": "E2", "name": "\textbf{Lymphocytes}", "type": "Structure"\}\}, \\
            \hspace*{2em} \{\{"id": "E3", "name": "\textbf{Desmoplastic stroma}", "type": "Structure"\}\}, \\
            
            \vspace{0.2em}
            \hspace*{2em} \textcolor{gray}{// 2. Pathological Phenotypes} \\
            \hspace*{2em} \{\{"id": "P1", "name": "\textbf{Solid growth pattern}", "type": "Phenotype"\}\}, \\
            \hspace*{2em} \{\{"id": "P2", "name": "\textbf{High mitotic activity}", "type": "Phenotype"\}\}, \\
            \hspace*{2em} \{\{"id": "P3", "name": "\textbf{Comedo necrosis}", "type": "Phenotype"\}\}, \\
            \hspace*{2em} \{\{"id": "P4", "name": "\textbf{Tumor Infiltrating Lymphocytes (TILs)}", "type": "Phenotype"\}\}, \\
            
            \vspace{0.2em}
            \hspace*{2em} \textcolor{gray}{// 3. Diagnostic Concepts} \\
            \hspace*{2em} \{\{"id": "D1", "name": "\textbf{Invasive Ductal Carcinoma}", "type": "Diagnosis"\}\}, \\
            \hspace*{2em} \{\{"id": "D2", "name": "\textbf{Nottingham Grade III}", "type": "Diagnosis"\}\} \\
            \hspace*{1em} ] \\
            \}\}
            % \vspace{0.5em}
            % ==========================================
            % SEPARATOR 1: Extraction -> Retrieval
            % ==========================================
            \begin{center}
                \scriptsize\color{gray} $\Downarrow$ Querying Graph with Multi-Domain Entities $\Downarrow$
            \end{center}
            % \vspace{1em}
            % ==========================================
            % PHASE 2: RETRIEVAL
            % ==========================================
            \normalfont\sffamily
            \textbf{--- Step 2: Retrieved Knowledge Paths ---} \\
            % \vspace{0.2em}
            \ttfamily\footnotesize
            
            \textbf{Path A (Architecture):} \\
            \hspace*{1em} [Solid growth pattern] \textcolor{gray}{--implies$\to$} [No tubules] \textcolor{gray}{--suggests$\to$} [High Grade IDC] \\
            
            \textbf{Path B (Proliferation):} \\
            \hspace*{1em} [High mitotic activity] \textcolor{gray}{--componentOf$\to$} [Nottingham Grading] \textcolor{gray}{--increases$\to$} [Tumor Aggressiveness] \\
            
            \textbf{Path C (Microenvironment):} \\
            \hspace*{1em} [Lymphocytes] \textcolor{gray}{--infiltrate$\to$} [Desmoplastic stroma] \textcolor{gray}{--constitutes$\to$} [TILs] \textcolor{gray}{--associatedWith$\to$} [High Grade/Medullary features] \\

            \textbf{Path D (DCIS Context):} \\
            \hspace*{1em} [Comedo necrosis] \textcolor{gray}{--indicates$\to$} [Rapid cell turnover] \textcolor{gray}{--markerOf$\to$} [High Grade DCIS]

            % \vspace{1em}
            % ==========================================
            % SEPARATOR 2: Retrieval -> Reasoning
            % ==========================================
            \begin{center}
                \scriptsize\color{gray} $\Downarrow$ Injecting Paths into Reasoning Engine $\Downarrow$
            \end{center}
            % \vspace{1em}

            % ==========================================
            % PHASE 3: REASONING
            % ==========================================
            \normalfont\sffamily
            \textbf{--- Step 3: LLM-Integrated Chain of Thought ---} \\
            {
            \ttfamily
            \textbf{\textcolor{keywordcolor}{SYSTEM}:} You are a Knowledge-Grounded Reasoning Assistant. Your task is to synthesize a diagnostic rationale based on the provided Knowledge Graph paths. \\
            \vspace{0.2em}
            \textbf{\textcolor{keywordcolor}{USER}:}  Synthesize these paths into a coherent diagnostic chain. Explicitly state how the morphological evidence supports the Final Diagnosis and excludes alternatives.
            }
            % \vspace{0.4em}
            \vspace{0.2em}

            \textbf{Reasoning Output:} \\
            {
            \linespread{1.1}\selectfont 
            \small \itshape 
            ``The diagnostic reasoning begins with the architecture: the tumor exhibits a \underline{Solid growth pattern} with no tubule formation, which is the primary criteria for classifying the lesion as \textbf{Invasive Ductal Carcinoma}. 
            
            Integrating the proliferation metrics, the \underline{High mitotic activity} serves as a critical biomarker that \underline{increases} the calculated grade, firmly establishing the diagnosis as \textbf{Nottingham Grade III}. 
            
            Crucially, the analysis extends to the tumor microenvironment. The presence of \underline{Lymphocytes} that \underline{infiltrate} the \underline{Desmoplastic stroma} constitutes significant \textbf{Tumor Infiltrating Lymphocytes (TILs)}. The combination of a solid growth pattern, high grade, and intense lymphocytic infiltrate raises the specific morphological subtype of \textbf{Carcinoma with Medullary Features}.
            Finally, the concurrent \underline{Comedo necrosis} confirms the presence of high-grade DCIS. 
            
            Conclusion: High-grade Invasive Ductal Carcinoma with prominent TILs."
            \par 
            }
            \vspace{0.4em}
        };
        
        % 标题 Tab
        \node[
            fill=framecolor,
            text=white,
            rounded corners={5pt},
            anchor=west,
            xshift=20pt
        ] at (box.north west) {
            \bfseries \sffamily \hspace{0.5em} Reasoning Generation Pipeline \hspace{0.5em}
        };
        
    \end{tikzpicture}
    
    \caption{\textbf{Advanced Reasoning Pipeline (Immune-Oncology Context).} This pipeline demonstrates the integration of standard histological grading with microenvironmental features. Step 1 extracts entities related to both tumor cells and the immune response. Step 2 retrieves paths connecting stromal lymphocytes to the concept of TILs, running parallel to standard grading paths. Step 3 synthesizes a diagnosis that captures the tumor's biological complexity.}
\label{ReasonPipeline4}
\end{figure}

% TCGA-A2-A0YL-01Z-00-DX1.69A438C7-B1E0-4990-B1E6-586C551DC79C
\begin{figure}[!p]
    \centering
    \begin{tikzpicture}
        % 主内容框
        \node[
            draw=framecolor,
            line width=1.5pt,
            rounded corners=8pt,
            inner sep=12pt,
            text width=16cm, 
            align=left,
            font=\small\sffamily,
            fill=codebg
        ] (box) {
            
            % ==========================================
            % PHASE 1: EXTRACTION
            % ==========================================
            {
            \ttfamily
            \textbf{\textcolor{keywordcolor}{SYSTEM}:} You are a Pathological Entity Extraction Engine. Extract terms strictly adhering to the schema: [\texttt{Physical\_Entity}, \texttt{Phenotype}, \texttt{Diagnosis}]. \\
            \vspace{0.2em}
            \textbf{\textcolor{keywordcolor}{USER}:} Analyze the following pathology report. Identify precise mention spans, map them to the schema types, and output the result in JSON format.
            }
            
            \vspace{0.3em}
            \textbf{--- Input Text (Excerpt) ---} \\
            \textit{"...cells are markedly \underline{discohesive} and arranged in \underline{linear cords} (one cell thick) composed of cells with moderate nuclear atypia and rare mitotic figures ... swirling around benign ducts creating a \underline{targetoid pattern} ..."}
            
            \vspace{0.3em}
            \textbf{--- Step 1: Model Output (Specific ILC Features) ---} \\

            \ttfamily 
            \{\{ \\
            \hspace*{1em} "extracted\_entities": [ \\
            \hspace*{2em} \textcolor{gray}{// 1. Physical Entities} \\
            \hspace*{2em} \{\{"id": "E1", "name": "\textbf{Linear cords}", "type": "Structure"\}\}, \\
            \hspace*{2em} \{\{"id": "E2", "name": "\textbf{Benign ducts}", "type": "Structure"\}\}, \\
            \hspace*{2em} \{\{"id": "E3", "name": "\textbf{Discohesive cells}", "type": "Structure"\}\}, \\
            \hspace*{2em} \{\{"id": "E4", "name": "\textbf{Fibrous stroma}", "type": "Structure"\}\}, \\
            
            \vspace{0.2em}
            \hspace*{2em} \textcolor{gray}{// 2. Pathological Phenotypes} \\
            \hspace*{2em} \{\{"id": "P1", "name": "\textbf{Indian filing}", "type": "Phenotype"\}\}, \\
            \hspace*{2em} \{\{"id": "P2", "name": "\textbf{Targetoid pattern}", "type": "Phenotype"\}\}, \\
            \hspace*{2em} \{\{"id": "P3", "name": "\textbf{Low mitotic index}", "type": "Phenotype"\}\}, \\
            
            \vspace{0.2em}
            \hspace*{2em} \textcolor{gray}{// 3. Diagnostic Concepts} \\
            \hspace*{2em} \{\{"id": "D1", "name": "\textbf{Invasive Lobular Carcinoma (ILC)}", "type": "Diagnosis"\}\}, \\
            \hspace*{2em} \{\{"id": "D2", "name": "\textbf{Invasive Ductal Carcinoma}", "type": "Diagnosis (Excluded)"\}\} \\
            \hspace*{1em} ] \\
            \}\}
            % \vspace{0.5em}
            % ==========================================
            % SEPARATOR 1: Extraction -> Retrieval
            % ==========================================
            \begin{center}
                \scriptsize\color{gray} $\Downarrow$ Querying Graph with Morphological Patterns $\Downarrow$
            \end{center}
            % \vspace{1em}
            % ==========================================
            % PHASE 2: RETRIEVAL
            % ==========================================
            \normalfont\sffamily
            \textbf{--- Step 2: Retrieved Knowledge Paths (Distinctive Morphology) ---} \\
            % \vspace{0.2em}
            \ttfamily\footnotesize
            
            \textbf{Path A (The "Classic" ILC Path):} \\
            \hspace*{1em} [Discohesive cells] \textcolor{gray}{--arrangeIn$\to$} [Linear cords] \textcolor{gray}{--knownAs$\to$} [Indian filing] \textcolor{gray}{--diagnosticOf$\to$} [ILC] \\
            
            \textbf{Path B (The "Targetoid" Path):} \\
            \hspace*{1em} [Tumor cells] \textcolor{gray}{--swirlAround$\to$} [Benign ducts] \textcolor{gray}{--forms$\to$} [Targetoid pattern] \textcolor{gray}{--characteristicOf$\to$} [ILC] \\
            
            \textbf{Path C (Differentiation):} \\
            \hspace*{1em} [Discohesive cells] \textcolor{gray}{--lacks$\to$} [Gland formation] \textcolor{gray}{--distinguishesFrom$\to$} [Invasive Ductal Carcinoma] \\

            \textbf{Path D (Grading Context):} \\
            \hspace*{1em} [Low mitotic index] \textcolor{gray}{--plus$\to$} [Moderate nuclear atypia] \textcolor{gray}{--resultsIn$\to$} [Nottingham Grade II]

            % \vspace{1em}
            % ==========================================
            % SEPARATOR 2: Retrieval -> Reasoning
            % ==========================================
            \begin{center}
                \scriptsize\color{gray} $\Downarrow$ Injecting Paths into Reasoning Engine $\Downarrow$
            \end{center}
            % \vspace{1em}

            % ==========================================
            % PHASE 3: REASONING
            % ==========================================
            \normalfont\sffamily
            \textbf{--- Step 3: LLM-Integrated Chain of Thought ---} \\
            {
            \ttfamily
            \textbf{\textcolor{keywordcolor}{SYSTEM}:} You are a Knowledge-Grounded Reasoning Assistant. Your task is to synthesize a diagnostic rationale based on the provided Knowledge Graph paths. \\
            \vspace{0.2em}
            \textbf{\textcolor{keywordcolor}{USER}:}  Synthesize these paths into a coherent diagnostic chain. Explicitly state how the morphological evidence supports the Final Diagnosis and excludes alternatives.
            }
            % \vspace{0.4em}
            \vspace{0.2em}

            \textbf{Reasoning Output:} \\
            {
            \linespread{1.1}\selectfont 
            \small \itshape 
            ``The primary diagnostic task is to interpret the unique growth pattern. 
            Unlike ductal carcinoma which typically forms glands, the retrieved entities show \underline{Discohesive cells} that fail to adhere to one another. 
            Instead, these cells arrange themselves into \underline{Linear cords} within the fibrous stroma. This specific phenomenon is morphologically identified as \textbf{Indian filing}, which is the pathognomonic feature of \textbf{Invasive Lobular Carcinoma (ILC)}.
            
            A secondary confirming feature is observed where tumor cells \underline{swirl around} preserved \underline{Benign ducts}, creating a \textbf{Targetoid pattern}. 

            Regarding the grade: ILC inherently \underline{lacks tubule formation} (Score 3). Combined with the observed \underline{Moderate nuclear atypia} (Score 2) and \underline{Low mitotic index} (Score 1), the aggregate score (6) classifies this as \textbf{Nottingham Grade II}.
            
            Conclusion: The architecture strictly confirms Invasive Lobular Carcinoma, Grade II."
            \par 
            }
            \vspace{0.4em}
        };
        
        % 标题 Tab
        \node[
            fill=framecolor,
            text=white,
            rounded corners={5pt},
            anchor=west,
            xshift=20pt
        ] at (box.north west) {
            \bfseries \sffamily \hspace{0.5em} Reasoning Generation Pipeline \hspace{0.5em}
        };
        
    \end{tikzpicture}
    
    \caption{\textbf{Distinctive Reasoning Pipeline (Lobular Carcinoma).} This pipeline highlights the specific morphological reasoning required for Lobular Carcinoma. Step 1 focuses on architectural entities like Linear cords and Discohesive cells. Step 2 retrieves paths that link these patterns like Indian filing and Targetoid directly to ILC, explicitly distinguishing it from Ductal Carcinoma . Step 3 synthesizes these unique patterns into a definitive diagnosis.}
\label{ReasonPipeline5}
\end{figure}

%TCGA-A7-A5ZV-01Z-00-DX1.21F2EA4A-4F31-43D6-A036-E20E326AF37E
\begin{figure}[!p]
    \centering
    \begin{tikzpicture}
        % 主内容框
        \node[
            draw=framecolor,
            line width=1.5pt,
            rounded corners=8pt,
            inner sep=12pt,
            text width=16cm, 
            align=left,
            font=\small\sffamily,
            fill=codebg
        ] (box) {
            
            % ==========================================
            % PHASE 1: EXTRACTION
            % ==========================================
            {
            \ttfamily
            \textbf{\textcolor{keywordcolor}{SYSTEM}:} You are a Pathological Entity Extraction Engine. Extract terms strictly adhering to the schema: [\texttt{Physical\_Entity}, \texttt{Phenotype}, \texttt{Diagnosis}]. \\
            \vspace{0.2em}
            \textbf{\textcolor{keywordcolor}{USER}:} Analyze the following pathology report. Identify precise mention spans, map them to the schema types, and output the result in JSON format.
            }
            
            \vspace{0.3em}
            \textbf{--- Input Text (Excerpt) ---} \\
            \textit{"...exhibits a \underline{pushing border}, expanding with a \underline{broad front} ... cells with marked pleomorphism grow in \underline{syncytial sheets} ... prominent \underline{lymphoplasmacytic infiltrate}..."}
            
            \vspace{0.3em}
            \textbf{--- Step 1: Model Output (Architectural Entities) ---} \\

            \ttfamily 
            \{\{ \\
            \hspace*{1em} "extracted\_entities": [ \\
            \hspace*{2em} \textcolor{gray}{// 1. Physical Entities} \\
            \hspace*{2em} \{\{"id": "E1", "name": "\textbf{Pushing border}", "type": "Structure"\}\}, \\
            \hspace*{2em} \{\{"id": "E2", "name": "\textbf{Syncytial sheets}", "type": "Structure"\}\}, \\
            \hspace*{2em} \{\{"id": "E3", "name": "\textbf{Broad front}", "type": "Structure"\}\}, \\
            
            \vspace{0.2em}
            \hspace*{2em} \textcolor{gray}{// 2. Pathological Phenotypes} \\
            \hspace*{2em} \{\{"id": "P1", "name": "\textbf{Circumscribed growth}", "type": "Phenotype"\}\}, \\
            \hspace*{2em} \{\{"id": "P2", "name": "\textbf{High nuclear grade}", "type": "Phenotype"\}\}, \\
            \hspace*{2em} \{\{"id": "P3", "name": "\textbf{Lymphoplasmacytic infiltrate}", "type": "Phenotype"\}\}, \\
            
            \vspace{0.2em}
            \hspace*{2em} \textcolor{gray}{// 3. Diagnostic Concepts} \\
            \hspace*{2em} \{\{"id": "D1", "name": "\textbf{Invasive Carcinoma (NST)}", "type": "Diagnosis"\}\}, \\
            \hspace*{2em} \{\{"id": "D2", "name": "\textbf{Medullary-like features}", "type": "Diagnosis"\}\} \\
            \hspace*{1em} ] \\
            \}\}
            % \vspace{0.5em}
            % ==========================================
            % SEPARATOR 1: Extraction -> Retrieval
            % ==========================================
            \begin{center}
                \scriptsize\color{gray} $\Downarrow$ Querying Graph with Architectural Patterns $\Downarrow$
            \end{center}
            % \vspace{1em}
            % ==========================================
            % PHASE 2: RETRIEVAL
            % ==========================================
            \normalfont\sffamily
            \textbf{--- Step 2: Retrieved Knowledge Paths (Contrast Logic) ---} \\
            % \vspace{0.2em}
            \ttfamily\footnotesize
            
            \textbf{Path A (The High-Grade Logic):} \\
            \hspace*{1em} [Syncytial sheets] \textcolor{gray}{--lack$\to$} [Tubules] \textcolor{gray}{--plus$\to$} [High Pleomorphism] \textcolor{gray}{--classifiesAs$\to$} [Grade III] \\
            
            \textbf{Path B (The Architectural Logic):} \\
            \hspace*{1em} [Broad front] \textcolor{gray}{--defines$\to$} [Pushing border] \textcolor{gray}{--contrastsWith$\to$} [Infiltrative/Stellate border] \\
            
            \textbf{Path C (The Specific Subtype Logic):} \\
            \hspace*{1em} [Pushing border] \textcolor{gray}{--combinedWith$\to$} [Lymphoplasmacytic infiltrate] \textcolor{gray}{--suggests$\to$} [Medullary-like features] \\

            \textbf{Path D (Prognostic Implication):} \\
            \hspace*{1em} [Circumscribed growth] \textcolor{gray}{--oftenCorrelatesWith$\to$} [Lower rate of Nodal Metastasis]

            % \vspace{1em}
            % ==========================================
            % SEPARATOR 2: Retrieval -> Reasoning
            % ==========================================
            \begin{center}
                \scriptsize\color{gray} $\Downarrow$ Injecting Paths into Reasoning Engine $\Downarrow$
            \end{center}
            % \vspace{1em}

            % ==========================================
            % PHASE 3: REASONING
            % ==========================================
            \normalfont\sffamily
            \textbf{--- Step 3: LLM-Integrated Chain of Thought ---} \\
            {
            \ttfamily
            \textbf{\textcolor{keywordcolor}{SYSTEM}:} You are a Knowledge-Grounded Reasoning Assistant. Your task is to synthesize a diagnostic rationale based on the provided Knowledge Graph paths. \\
            \vspace{0.2em}
            \textbf{\textcolor{keywordcolor}{USER}:}  Synthesize these paths into a coherent diagnostic chain. Explicitly state how the morphological evidence supports the Final Diagnosis and excludes alternatives.
            }
            % \vspace{0.4em}
            \vspace{0.2em}

            \textbf{Reasoning Output:} \\
            {
            \linespread{1.1}\selectfont 
            \small \itshape 
            ``We start by evaluating the tumor grade. The cells grow in \underline{Syncytial sheets} with high marked nuclear pleomorphism, definitively establishing this as a \textbf{High-Grade Carcinoma}. Typically, high-grade tumors exhibit aggressive, stellate infiltration.
            
            However, this case presents a distinct architectural deviation. The tumor advances with a \underline{Broad front}, creating a \underline{Pushing border} rather than an infiltrative one. This \underline{Circumscribed growth} pattern is crucial for differential diagnosis.
            
            When we combine the \underline{Pushing border} with the prominent \underline{Lymphoplasmacytic infiltrate}, the pattern points towards \textbf{Invasive Carcinoma with Medullary-like features}. 
            
            This specific morphology helps explain the clinical picture, as such tumors, despite being high-grade, often have a more favorable nodal status compared to classic spiculated high-grade IDC.
            
            Conclusion: High-Grade Invasive Carcinoma with pushing borders."
            \par 
            }
            \vspace{0.4em}
        };
        
        % 标题 Tab
        \node[
            fill=framecolor,
            text=white,
            rounded corners={5pt},
            anchor=west,
            xshift=20pt
        ] at (box.north west) {
            \bfseries \sffamily \hspace{0.5em} Reasoning Generation Pipeline \hspace{0.5em}
        };
        
    \end{tikzpicture}
    
    \caption{\textbf{Nuanced Reasoning Pipeline (Pushing Borders).} Step 1 extracts the contradictory features (Syncytial sheets vs. Pushing border). Step 2 retrieves paths that reconcile these features under the concept of Medullary-like or Circumscribed high-grade carcinoma. Step 3 synthesizes a sophisticated diagnosis that explains why a high-grade tumor might have clear margins and negative nodes.}
\label{ReasonPipeline6}
\end{figure}

% TCGA-AO-A03V-01Z-00-DX1.52EBCB72-0C65-4E67-B9BB-DA15494327DE
\begin{figure}[!p]
    \centering
    \begin{tikzpicture}
        % 主内容框
        \node[
            draw=framecolor,
            line width=1.5pt,
            rounded corners=8pt,
            inner sep=12pt,
            text width=16cm, 
            align=left,
            font=\small\sffamily,
            fill=codebg
        ] (box) {
            
            % ==========================================
            % PHASE 1: EXTRACTION
            % ==========================================
            {
            \ttfamily
            \textbf{\textcolor{keywordcolor}{SYSTEM}:} You are a Pathological Entity Extraction Engine. Extract terms strictly adhering to the schema: [\texttt{Physical\_Entity}, \texttt{Phenotype}, \texttt{Diagnosis}]. \\
            \vspace{0.2em}
            \textbf{\textcolor{keywordcolor}{USER}:} Analyze the following pathology report. Identify precise mention spans, map them to the schema types, and output the result in JSON format.
            }
            
            \vspace{0.3em}
            \textbf{--- Input Text (Excerpt) ---} \\
            \textit{"...mixture of distinct \underline{glandular tubules} with moderate nuclear pleomorphism and infrequent mitotic figures ... \underline{punched-out spaces} consistent with \underline{cribriform pattern} ... \underline{basophilic deposits} representing \underline{microcalcifications}..."}
            
            \vspace{0.3em}
            \textbf{--- Step 1: Model Output (Specific Morphological Features) ---} \\

            \ttfamily 
            \{\{ \\
            \hspace*{1em} "extracted\_entities": [ \\
            \hspace*{2em} \textcolor{gray}{// 1. Physical Entities} \\
            \hspace*{2em} \{\{"id": "E1", "name": "\textbf{Glandular tubules}", "type": "Structure"\}\}, \\
            \hspace*{2em} \{\{"id": "E2", "name": "\textbf{Basophilic deposits}", "type": "Structure"\}\}, \\
            \hspace*{2em} \{\{"id": "E3", "name": "\textbf{Punched-out spaces}", "type": "Structure"\}\}, \\
            
            \vspace{0.2em}
            \hspace*{2em} \textcolor{gray}{// 2. Pathological Phenotypes} \\
            \hspace*{2em} \{\{"id": "P1", "name": "\textbf{Partial differentiation}", "type": "Phenotype"\}\}, \\
            \hspace*{2em} \{\{"id": "P2", "name": "\textbf{Cribriform pattern}", "type": "Phenotype"\}\}, \\
            \hspace*{2em} \{\{"id": "P3", "name": "\textbf{Microcalcifications}", "type": "Phenotype"\}\}, \\
            
            \vspace{0.2em}
            \hspace*{2em} \textcolor{gray}{// 3. Diagnostic Concepts} \\
            \hspace*{2em} \{\{"id": "D1", "name": "\textbf{Invasive Ductal Carcinoma}", "type": "Diagnosis"\}\}, \\
            \hspace*{2em} \{\{"id": "D2", "name": "\textbf{Ductal Carcinoma In Situ (DCIS)}", "type": "Diagnosis"\}\} \\
            \hspace*{1em} ] \\
            \}\}
            % \vspace{0.5em}
            % ==========================================
            % SEPARATOR 1: Extraction -> Retrieval
            % ==========================================
            \begin{center}
                \scriptsize\color{gray} $\Downarrow$ Querying Graph with Morphological Entities $\Downarrow$
            \end{center}
            % \vspace{1em}
            % ==========================================
            % PHASE 2: RETRIEVAL
            % ==========================================
            \normalfont\sffamily
            \textbf{--- Step 2: Retrieved Knowledge Paths (Logic Graph) ---} \\
            % \vspace{0.2em}
            \ttfamily\footnotesize
            
            \textbf{Path A (Invasive Architecture):} \\
            \hspace*{1em} [Glandular tubules] \textcolor{gray}{--presentButLimitedTo$\to$} [10-75\%] \textcolor{gray}{--contributesTo$\to$} [Grade II (Moderate)] \\
            
            \textbf{Path B (DCIS Morphology):} \\
            \hspace*{1em} [Punched-out spaces] \textcolor{gray}{--resemble$\to$} [Sieve-like structure] \textcolor{gray}{--characteristicOf$\to$} [Cribriform DCIS] \\
            
            \textbf{Path C (Calcification Context):} \\
            \hspace*{1em} [Basophilic deposits] \textcolor{gray}{--locatedIn$\to$} [Duct lumen] \textcolor{gray}{--identifiesAs$\to$} [Microcalcifications] \textcolor{gray}{--associatedWith$\to$} [DCIS] \\

            \textbf{Path D (Diagnostic Synthesis):} \\
            \hspace*{1em} [Invasive Ductal Carcinoma] \textcolor{gray}{--coexistsWith$\to$} [DCIS] \textcolor{gray}{--implies$\to$} [Local Origin]

            % \vspace{1em}
            % ==========================================
            % SEPARATOR 2: Retrieval -> Reasoning
            % ==========================================
            \begin{center}
                \scriptsize\color{gray} $\Downarrow$ Injecting Paths into Reasoning Engine $\Downarrow$
            \end{center}
            % \vspace{1em}

            % ==========================================
            % PHASE 3: REASONING
            % ==========================================
            \normalfont\sffamily
            \textbf{--- Step 3: LLM-Integrated Chain of Thought ---} \\
            {
            \ttfamily
            \textbf{\textcolor{keywordcolor}{SYSTEM}:} You are a Knowledge-Grounded Reasoning Assistant. Your task is to synthesize a diagnostic rationale based on the provided Knowledge Graph paths. \\
            \vspace{0.2em}
            \textbf{\textcolor{keywordcolor}{USER}:}  Synthesize these paths into a coherent diagnostic chain. Explicitly state how the morphological evidence supports the Final Diagnosis and excludes alternatives.
            }
            % \vspace{0.4em}
            \vspace{0.2em}

            \textbf{Reasoning Output:} \\
            {
            \linespread{1.1}\selectfont 
            \small \itshape 
            ``The diagnostic evaluation starts with the invasive component. 
            The diagnostic evaluation of the invasive component reveals the presence of \underline{Glandular tubules}, indicating the tumor retains some ability to differentiate. However, as this formation is partial (10-75\%), it is assigned a Score of 2. Additionally, the cells exhibit moderate nuclear pleomorphism (Score 2) and intermediate mitotic activity (Score 2). The summation of these values results in a total score of 6, strictly categorizing the tumor as \textbf{Nottingham Grade II (Moderately Differentiated)}.
            
            A crucial associated finding is observed in the adjacent ducts. The tumor cells form \underline{Punched-out spaces} creating a sieve-like appearance, which is the morphological definition of \textbf{Cribriform DCIS}. 
            
            Furthermore, \underline{Basophilic deposits} are identified within these lumens. Morphologically, these represent \textbf{Microcalcifications}, a critical feature often correlating with mammographic findings and confirming the diagnosis of the in-situ component.
            
            Conclusion: Moderately differentiated (Grade II) Invasive Ductal Carcinoma with calcified Cribriform DCIS."
            \par 
            }
            \vspace{0.4em}
        };
        
        % 标题 Tab
        \node[
            fill=framecolor,
            text=white,
            rounded corners={5pt},
            anchor=west,
            xshift=20pt
        ] at (box.north west) {
            \bfseries \sffamily \hspace{0.5em} Reasoning Generation Pipeline \hspace{0.5em}
        };
        
    \end{tikzpicture}
    
    \caption{\textbf{Integrative Reasoning Pipeline (Grade II IDC with DCIS).} This pipeline addresses the diagnosis of a challenging tumor subtype. Step 1 extracts features of lobular architecture and high-grade nuclear atypia. Step 2 retrieves paths that link the single-file pattern to lobular lineage and connects the nuclear features to high-grade deviation. Step 3 synthesizes these distinct elements to construct a definitive diagnosis of the pleomorphic variant.}
\label{ReasonPipeline7}
\end{figure}

% TCGA-AO-A12G-01Z-00-DX1.1C37E1FE-CF36-4570-A864-3813B8ADBA36
\begin{figure}[!p]
    \centering
    \begin{tikzpicture}
        % 主内容框
        \node[
            draw=framecolor,
            line width=1.5pt,
            rounded corners=8pt,
            inner sep=12pt,
            text width=16cm, 
            align=left,
            font=\small\sffamily,
            fill=codebg
        ] (box) {
            
            % ==========================================
            % PHASE 1: EXTRACTION
            % ==========================================
            {
            \ttfamily
            \textbf{\textcolor{keywordcolor}{SYSTEM}:} You are a Pathological Entity Extraction Engine. Extract terms strictly adhering to the schema: [\texttt{Physical\_Entity}, \texttt{Phenotype}, \texttt{Diagnosis}]. \\
            \vspace{0.2em}
            \textbf{\textcolor{keywordcolor}{USER}:} Analyze the following pathology report. Identify precise mention spans, map them to the schema types, and output the result in JSON format.
            }
            
            \vspace{0.3em}
            \textbf{--- Input Text (Excerpt) ---} \\
            \textit{"...cells are \underline{discohesive} and arranged in \underline{Indian files} ... exhibit \underline{marked nuclear pleomorphism} with \underline{prominent nucleoli} ... abundant \underline{eosinophilic cytoplasm} adjacent to foci of \underline{Lobular Carcinoma In Situ (LCIS)} ..."}
            
            \vspace{0.3em}
            \textbf{--- Step 1: Model Output (Hybrid Features) ---} \\

            \ttfamily 
            \{\{ \\
            \hspace*{1em} "extracted\_entities": [ \\
            \hspace*{2em} \textcolor{gray}{// 1. Physical Entities} \\
            \hspace*{2em} \{\{"id": "E1", "name": "\textbf{Discohesive cells}", "type": "Structure"\}\}, \\
            \hspace*{2em} \{\{"id": "E2", "name": "\textbf{Indian files}", "type": "Structure"\}\}, \\
            \hspace*{2em} \{\{"id": "E3", "name": "\textbf{Prominent nucleoli}", "type": "Structure"\}\}, \\
            \hspace*{2em} \{\{"id": "E4", "name": "\textbf{LCIS}", "type": "Structure"\}\}, \\
            
            \vspace{0.2em}
            \hspace*{2em} \textcolor{gray}{// 2. Pathological Phenotypes} \\
            \hspace*{2em} \{\{"id": "P1", "name": "\textbf{Infiltrative growth}", "type": "Phenotype"\}\}, \\
            \hspace*{2em} \{\{"id": "P2", "name": "\textbf{High nuclear grade}", "type": "Phenotype"\}\}, \\
            \hspace*{2em} \{\{"id": "P3", "name": "\textbf{Eosinophilic features}", "type": "Phenotype"\}\}, \\
            
            \vspace{0.2em}
            \hspace*{2em} \textcolor{gray}{// 3. Diagnostic Concepts} \\
            \hspace*{2em} \{\{"id": "D1", "name": "\textbf{Invasive Lobular Carcinoma}", "type": "Diagnosis"\}\}, \\
            \hspace*{2em} \{\{"id": "D2", "name": "\textbf{Pleomorphic Type}", "type": "Diagnosis"\}\} \\
            \hspace*{1em} ] \\
            \}\}
            % \vspace{0.5em}
            % ==========================================
            % SEPARATOR 1: Extraction -> Retrieval
            % ==========================================
            \begin{center}
                \scriptsize $\Downarrow$ Querying Graph with Hybrid Attributes $\Downarrow$
            \end{center}
            % \vspace{1em}
            % ==========================================
            % PHASE 2: RETRIEVAL
            % ==========================================
            \normalfont\sffamily
            \textbf{--- Step 2: Retrieved Knowledge Paths (Subtype Logic) ---} \\
            % \vspace{0.2em}
            \ttfamily\footnotesize
            
            \textbf{Path A (Architectural Signature):} \\
            \hspace*{1em} [Discohesive cells] \textcolor{gray}{--form$\to$} [Indian files] \textcolor{gray}{--morphologicalProxyFor$\to$} [E-cadherin loss] \textcolor{gray}{--indicates$\to$} [ILC] \\
            
            \textbf{Path B (Cytological Deviation):} \\
            \hspace*{1em} [Prominent nucleoli] \textcolor{gray}{--defines$\to$} [High nuclear grade] \textcolor{gray}{--atypicalFor$\to$} [Classic ILC] \\
            
            \textbf{Path C (Diagnostic Synthesis):} \\
            \hspace*{1em} [ILC Architecture] \textcolor{gray}{--combinedWith$\to$} [High nuclear grade] \textcolor{gray}{--classifiesAs$\to$} [Pleomorphic ILC] \\

            \textbf{Path D (Origin Confirmation):} \\
            \hspace*{1em} [LCIS] \textcolor{gray}{--precursorOf$\to$} [Invasive Lobular Carcinoma]

            \vspace{1em}
            % ==========================================
            % SEPARATOR 2: Retrieval -> Reasoning
            % ==========================================
            \begin{center}
                \scriptsize\color{gray} $\Downarrow$ Injecting Paths into Reasoning Engine $\Downarrow$
            \end{center}
            % \vspace{1em}

            % ==========================================
            % PHASE 3: REASONING
            % ==========================================
            \normalfont\sffamily
            \textbf{--- Step 3: LLM-Integrated Chain of Thought ---} \\
            {
            \ttfamily
            \textbf{\textcolor{keywordcolor}{SYSTEM}:} You are a Knowledge-Grounded Reasoning Assistant. Your task is to synthesize a diagnostic rationale based on the provided Knowledge Graph paths. \\
            \vspace{0.2em}
            \textbf{\textcolor{keywordcolor}{USER}:}  Synthesize these paths into a coherent diagnostic chain. Explicitly state how the morphological evidence supports the Final Diagnosis and excludes alternatives.
            }
            % \vspace{0.4em}
            \vspace{0.2em}

            \textbf{Reasoning Output:} \\
            {
            \linespread{1.1}\selectfont 
            \small \itshape 
            ``The reasoning requires integrating two distinct sets of features.
            First, structurally, the tumor cells are \underline{Discohesive} and arrange in \underline{Indian files}. This pattern is the definitive morphological proxy for E-cadherin loss, establishing the primary diagnosis as \textbf{Invasive Lobular Carcinoma (ILC)}.
            
            Second, the nuclei present with \underline{Prominent nucleoli} and \underline{High nuclear grade}. This is \underline{atypical for} Classic ILC, which usually consists of small, uniform cells.
            
            When the architectural framework of ILC is \underline{combined with} high-grade cytological features (and often \underline{Eosinophilic features}), the specific subtype is classified as \textbf{Pleomorphic Invasive Lobular Carcinoma}. The presence of adjacent \underline{LCIS} further validates the lobular lineage.
            
            Conclusion: Invasive Lobular Carcinoma, Pleomorphic Type."
            \par 
            }
            \vspace{0.4em}
        };
        
        % 标题 Tab
        \node[
            fill=framecolor,
            text=white,
            rounded corners={5pt},
            anchor=west,
            xshift=20pt
        ] at (box.north west) {
            \bfseries \sffamily \hspace{0.5em} Reasoning Generation Pipeline \hspace{0.5em}
        };
        
    \end{tikzpicture}
    
    \caption{\textbf{Hybrid Reasoning Pipeline (Pleomorphic Lobular Carcinoma).} This pipeline addresses a challenging subtype. Step 1 extracts features that seem contradictory: "Lobular" architecture (Indian files) vs. "Ductal-like" high grade (prominent nucleoli). Step 2 retrieves paths that explain this specific combination. Path A confirms the Lobular family, Path B notes the high-grade deviation, and Path C synthesizes them into the specific "Pleomorphic" subtype diagnosis.}
\label{ReasonPipeline_last}
\end{figure}

\subsection{Mask Trajectory Sampling}
\label{subsec:trajectory}

To align the training objective with the autoregressive nature of clinical reasoning, we implement a Mask Trajectory Sampling strategy. Formally, given a complete ground-truth reasoning chain $R=[s_1, \dots, s_L]$, we construct an augmented dataset $\mathcal{D}_{\text{aug}}$ by randomly sampling truncation points $m$ uniformly from $[1, L]$. For each augmented instance, the model is conditioned on the partial history $s_{1:m-1}$ and optimized to generate the remaining trajectory $s_{m:L}$.
Figures \ref{Augmentation3}--\ref{Augmentation2} provide concrete illustrations of this process using a multi-feature pathology case. As depicted in Figure \ref{Augmentation3}, a single coherent chain involving histological subtyping, grading, and vascular invasion is sliced into distinct training samples. This exposes the model to diverse reasoning states, from the initial identification of architectural patterns (Sample A) to the intermediate deduction of risk features like LVI (Sample B), and finally to the synthesis of the diagnostic conclusion (Sample C). By scaling the corpus with these variable-context instances, we ensure the model learns to robustly recover logic flow from any intermediate state, preventing reliance on fixed template patterns.

%% mask strategy

\begin{figure}[!p]
    \centering
    \begin{tikzpicture}
        % 主内容框
        \node[
            draw=framecolor,
            line width=1.5pt,
            rounded corners=8pt,
            inner sep=12pt,
            text width=16cm,
            align=left,
            font=\small\sffamily,
            fill=codebg
        ] (box) {
            
            % ==========================================
            % PHASE 1: ORIGINAL DATA (ANCHOR)
            % ==========================================
            {
            \ttfamily
            \textbf{\textcolor{keywordcolor}{SOURCE DATA}:} PathReasoner Dataset ($\mathcal{D}$). \\
            \textbf{\textcolor{keywordcolor}{STRATEGY}:} Trajectory Augmentation (Masking).
            }
            
            \vspace{0.5em}
            \textbf{--- Reference: Complete Reasoning Chain ($R$) ---} \\
            \textit{\footnotesize The full sequence involves main diagnosis, grading, and associated features.}
            \vspace{0.3em}
            
            {
            \linespread{1.05}\selectfont
            \itshape
            ``The histological architecture presents primarily as \underline{Solid sheets} with a conspicuous \underline{Lack of tubules}, which structurally characterizes the lesion as \textbf{Invasive Ductal Carcinoma}. ($s_1$)
            Simultaneously, the nuclear features show marked \underline{Pleomorphism} and \underline{Numerous mitoses}, establishing \underline{Nuclear Grade 3} and \underline{Mitotic Grade 3}, which drives the final classification to \textbf{Nottingham Grade III}. ($s_2$)
            Two critical associated features are identified via specific pathways: First, the presence of \underline{Central necrosis} within the ducts strictly \underline{defines} the \textbf{Comedo pattern}, indicating an associated High-Grade DCIS component. ($s_3$)
            Second, and most critically for prognosis, tumor clusters are located \underline{inside} \underline{Vascular spaces} that are \underline{lined by} \underline{Endothelium}. This specific anatomical relationship \underline{confirms} the diagnosis of \textbf{Lymphovascular Invasion (LVI)}. ($s_4$)
            Final Conclusion: High-grade Invasive Ductal Carcinoma with Comedo-DCIS and LVI. ($a$)"
            \par
            }

            \vspace{0.8em}
            % ==========================================
            % SEPARATOR
            % ==========================================
            \begin{center}
                \scriptsize\color{gray} $\Downarrow$ Generative Slicing: Creating $\mathcal{D}_{\text{aug}}$ Samples $\Downarrow$
            \end{center}
            \vspace{0.5em}
            
            % ==========================================
            % SAMPLE 1: Early Stage
            % ==========================================
            \normalfont\sffamily
            \textbf{--- Sample A: Grade \& Subtype Prediction ($m=2$) ---} \\
            
            \begin{quote}
                \ttfamily\footnotesize
                \textbf{\textcolor{keywordcolor}{INPUT CONTEXT} ($s_1$):} \\
                \normalfont\itshape\color{black}
                "The histological architecture presents primarily as \underline{Solid sheets} ... characterizes the lesion as \textbf{Invasive Ductal Carcinoma}."
                
                \vspace{0.2em}
                \ttfamily\footnotesize
                \textbf{\textcolor{keywordcolor}{TARGET LABEL} ($s_{2:L}, a$):} \\
                \normalfont\itshape\color{black}
                "Simultaneously, the nuclear features show marked \underline{Pleomorphism}... [Grading Logic] ... [Associated Features] ... Final Conclusion: High-grade Invasive Ductal Carcinoma with Comedo-DCIS and LVI."
            \end{quote}

            % ==========================================
            % SAMPLE 2: Mid Stage
            % ==========================================
            \vspace{0.2em}
            \normalfont\sffamily
            \textbf{--- Sample B: Risk Feature Identification ($m=4$) ---} \\
            \textit{\footnotesize The model sees the grading and DCIS, and must identify the critical LVI feature.}
            \vspace{0.2em}
            
            \begin{quote}
                \ttfamily\footnotesize
                \textbf{\textcolor{keywordcolor}{INPUT CONTEXT} ($s_{1:3}$):} \\
                \normalfont\itshape\color{black}
                "...indicating an associated High-Grade DCIS component." \textcolor{gray}{(End of Comedo DCIS logic)}
                
                \vspace{0.2em}
                \ttfamily\footnotesize
                \textbf{\textcolor{keywordcolor}{TARGET LABEL} ($s_{4:L}, a$):} \\
                \normalfont\itshape\color{black}
                "Second, and most critically for prognosis, tumor clusters are located \underline{inside} \underline{Vascular spaces} that are \underline{lined by} \underline{Endothelium}. This specific anatomical relationship \underline{confirms} the diagnosis of \textbf{Lymphovascular Invasion (LVI)}. Final Conclusion: High-grade Invasive Ductal Carcinoma with Comedo-DCIS and LVI."
            \end{quote}

            % ==========================================
            % SAMPLE 3: Late Stage
            % ==========================================
            \vspace{0.2em}
            \normalfont\sffamily
            \textbf{--- Sample C: Final Synthesis ($m=5$) ---} \\
            
            \begin{quote}
                \ttfamily\footnotesize
                \textbf{\textcolor{keywordcolor}{INPUT CONTEXT} ($s_{1:4}$):} \\
                \normalfont\itshape\color{black}
                "...This specific anatomical relationship \underline{confirms} the diagnosis of \textbf{Lymphovascular Invasion (LVI)}."
                
                \vspace{0.2em}
                \ttfamily\footnotesize
                \textbf{\textcolor{keywordcolor}{TARGET LABEL} ($a$):} \\
                \normalfont\itshape\color{black}
                "Final Conclusion: High-grade Invasive Ductal Carcinoma with Comedo-DCIS and LVI."
            \end{quote}
            
            \vspace{0.2em}
        };
        
        % 标题 Tab
        \node[
            fill=framecolor,
            text=white,
            rounded corners={5pt},
            anchor=west,
            xshift=20pt
        ] at (box.north west) {
            \bfseries \sffamily \hspace{0.5em} Trajectory Augmentation Example I
            \hspace{0.5em}
        };
        
    \end{tikzpicture}
    
    \caption{\textbf{Augmented Samples for Case Study I.} The process starts with a complete Golden Reason Chain ($R$). The augmentation strategy creates multiple training instances by truncating $R$ at random steps $m$. The model is provided with the partial history (Input Context) and trained to recover the remaining logic (Target Label), effectively teaching it to handle both early-stage generation and late-stage differential deduction.}
\label{Augmentation3}
\end{figure}

\begin{figure}[!p]
    \centering
    \begin{tikzpicture}
        % 主内容框
        \node[
            draw=framecolor,
            line width=1.5pt,
            rounded corners=8pt,
            inner sep=12pt,
            text width=16cm,
            align=left,
            font=\small\sffamily,
            fill=codebg
        ] (box) {
            
            % ==========================================
            % PHASE 1: ORIGINAL DATA (ANCHOR)
            % ==========================================
            {
            \ttfamily
            \textbf{\textcolor{keywordcolor}{SOURCE DATA}:} PathReasoner Dataset ($\mathcal{D}$). \\
            \textbf{\textcolor{keywordcolor}{STRATEGY}:} Trajectory Augmentation (Masking).
            }
            
            \vspace{0.5em}
            \textbf{--- Reference: Complete Reasoning Chain ($R$) ---} \\
            \textit{\footnotesize The full sequence before augmentation ($L=5$ steps + Answer).}
            \vspace{0.3em}
            
            {
            \linespread{1.05}\selectfont
            \itshape
            ``Histologically, the lesion is localized within the \underline{Main Bronchus}, where the architectural relationship between the surface epithelium and the \underline{adjacent} \underline{Basement Membrane} is scrutinized. ($s_1$)
            A defining pathological event is observed here: the \underline{Basement Membrane} serves as the direct \underline{site of} \textbf{Invasion} by neoplastic cells. ($s_2$)
            This breach is clinically significant because such invasion is a \underline{key feature} characteristic of \underline{Squamous Cell Carcinoma}. ($s_3$)
            For differential diagnosis, we distinguish this phenotype from \underline{Adenocarcinoma}. ($s_4$)
            The logic relies on morphological patterns: whereas \underline{Adenocarcinoma} predictably \underline{manifests} \underline{Glandular Structures}, the definition of \underline{Squamous Cell Carcinoma} explicitly \underline{excludes} them. ($s_5$)
            Therefore, the final diagnosis answer is squamous cell carcinoma. ($a$)"
            \par
            }

            \vspace{0.8em}
            % ==========================================
            % SEPARATOR: Definition -> Examples
            % ==========================================
            \begin{center}
                \scriptsize\color{gray} $\Downarrow$ Generative Slicing: Creating $\mathcal{D}_{\text{aug}}$ Samples $\Downarrow$
            \end{center}
            \vspace{0.5em}
            
            % ==========================================
            % SAMPLE 1: Early Stage
            % ==========================================
            \normalfont\sffamily
            \textbf{--- Sample A: Early-Stage Truncation ($m=2$) ---} \\
            
            \begin{quote}
                \ttfamily\footnotesize
                \textbf{\textcolor{keywordcolor}{INPUT CONTEXT} ($s_1$):} \\
                \normalfont\itshape\color{black}
                "Histologically, the lesion is localized within the \underline{Main Bronchus}..." \textcolor{gray}{(end of step 1)}
                
                \vspace{0.2em}
                \ttfamily\footnotesize
                \textbf{\textcolor{keywordcolor}{TARGET LABEL} ($s_{2:L}, a$):} \\
                \normalfont\itshape\color{black}
                "A defining pathological event is observed here: the \underline{Basement Membrane} serves as the direct \underline{site of} \textbf{Invasion} ... [Full remaining logic] ... Answer is squamous cell carcinoma."
            \end{quote}

            % ==========================================
            % SAMPLE 2: Mid Stage
            % ==========================================
            \vspace{0.2em}
            \normalfont\sffamily
            \textbf{--- Sample B: Differential Phase Truncation ($m=4$) ---} \\
            
            \begin{quote}
                \ttfamily\footnotesize
                \textbf{\textcolor{keywordcolor}{INPUT CONTEXT} ($s_{1:3}$):} \\
                \normalfont\itshape\color{black}
                "...[History]... such invasion is a \underline{key feature} characteristic of \underline{Squamous Cell Carcinoma}." \textcolor{gray}{(Model sees invasion evidence)}
                
                \vspace{0.2em}
                \ttfamily\footnotesize
                \textbf{\textcolor{keywordcolor}{TARGET LABEL} ($s_{4:L}, a$):} \\
                \normalfont\itshape\color{black}
                "For differential diagnosis, we distinguish this phenotype from \underline{Adenocarcinoma}. ... [Exclusion logic] ... Answer is squamous cell carcinoma."
            \end{quote}

            % ==========================================
            % SAMPLE 3: Late Stage
            % ==========================================
            \vspace{0.2em}
            \normalfont\sffamily
            \textbf{--- Sample C: Conclusion Phase Truncation ($m=5$) ---} \\
            
            \begin{quote}
                \ttfamily\footnotesize
                \textbf{\textcolor{keywordcolor}{INPUT CONTEXT} ($s_{1:4}$):} \\
                \normalfont\itshape\color{black}
                "...For differential diagnosis, we distinguish this phenotype from \underline{Adenocarcinoma}."
                
                \vspace{0.2em}
                \ttfamily\footnotesize
                \textbf{\textcolor{keywordcolor}{TARGET LABEL} ($s_{5}, a$):} \\
                \normalfont\itshape\color{black}
                "The logic relies on morphological patterns: whereas \underline{Adenocarcinoma} predictably \underline{manifests} \underline{Glandular Structures}, the definition of \underline{Squamous Cell Carcinoma} explicitly \underline{excludes} them. Therefore, the final diagnosis answer is squamous cell carcinoma."
            \end{quote}
            
            \vspace{0.2em}
        };
        
        % 标题 Tab
        \node[
            fill=framecolor,
            text=white,
            rounded corners={5pt},
            anchor=west,
            xshift=20pt
        ] at (box.north west) {
            \bfseries \sffamily \hspace{0.5em} Trajectory Augmentation Example II \hspace{0.5em}
        };
        
    \end{tikzpicture}
    
    \caption{\textbf{Augmented Samples for Case Study II.} The process starts with a complete Golden Reason Chain ($R$). The augmentation strategy creates multiple training instances by truncating $R$ at random steps $m$. The model is provided with the partial history (Input Context) and trained to recover the remaining logic (Target Label), effectively teaching it to handle both early-stage generation and late-stage differential deduction.}
\label{Augmentation1}
\end{figure}

\begin{figure}[!p]
    \centering
    \begin{tikzpicture}
        % 主内容框
        \node[
            draw=framecolor,
            line width=1.5pt,
            rounded corners=8pt,
            inner sep=12pt,
            text width=16cm,
            align=left,
            font=\small\sffamily,
            fill=codebg
        ] (box) {
            
            % ==========================================
            % PHASE 1: ORIGINAL DATA (ANCHOR)
            % ==========================================
            {
            \ttfamily
            \textbf{\textcolor{keywordcolor}{SOURCE DATA}:} PathReasoner Dataset ($\mathcal{D}$). \\
            \textbf{\textcolor{keywordcolor}{STRATEGY}:} Trajectory Augmentation (Masking).
            }
            
            \vspace{0.5em}
            \textbf{--- Reference: Complete Reasoning Chain ($R$) ---} \\
            \textit{\footnotesize The full sequence before augmentation. (Markers $s_1$-$s_5$ indicate logical steps).}
            \vspace{0.3em}
            
            {
            \linespread{1.05}\selectfont
            \itshape
            % -----------------------------------------------------------
            % [这里填入完整的推理文本]
            % 建议手动添加 ($s_1$), ($s_2$) 等标记，方便读者对应
            % -----------------------------------------------------------
            ``[Step 1: Initial Observation] The patient presents with \underline{Entity A} and \underline{Entity B}, suggesting an inflammatory process. ($s_1$)
            [Step 2: Evidence] A crucial finding is \textbf{Feature X} located in the \underline{Anatomical Region Y}. ($s_2$)
            [Step 3: Diagnosis] This pattern is highly indicative of \underline{Disease Main}, particularly given the presence of Feature X. ($s_3$)
            [Step 4: Differential/Exclusion] We must distinguish this from \underline{Disease Alternative}. Unlike Disease Main, Disease Alternative typically lacks \textbf{Feature X} and presents with Feature Z. ($s_4$)
            [Step 5: Conclusion] Given the absence of Feature Z, the diagnosis is confirmed as \underline{Disease Main}. ($a$)"
            \par
            }

            \vspace{0.8em}
            % ==========================================
            % SEPARATOR
            % ==========================================
            \begin{center}
                \scriptsize\color{gray} $\Downarrow$ Generative Slicing: Creating $\mathcal{D}_{\text{aug}}$ Samples $\Downarrow$
            \end{center}
            \vspace{0.5em}
            
            % ==========================================
            % SAMPLE 1: Early Stage (m=2)
            % ==========================================
            \normalfont\sffamily
            \textbf{--- Sample A: Early-Stage Truncation ($m=2$) ---} \\
            
            \begin{quote}
                \ttfamily\footnotesize
                \textbf{\textcolor{keywordcolor}{INPUT CONTEXT} ($s_1$):} \\
                \normalfont\itshape\color{black}
                % [填入 Step 1]
                "[Step 1 Text] The patient presents with \underline{Entity A}..." \textcolor{gray}{(end of step 1)}
                
                \vspace{0.2em}
                \ttfamily\footnotesize
                \textbf{\textcolor{keywordcolor}{TARGET LABEL} ($s_{2:L}, a$):} \\
                \normalfont\itshape\color{black}
                % [填入 Step 2 到最后]
                "[Step 2 Text] A crucial finding is... [Full remaining logic] ... diagnosis is confirmed as Disease Main."
            \end{quote}

            % ==========================================
            % SAMPLE 2: Mid Stage (m=4)
            % ==========================================
            \vspace{0.2em}
            \normalfont\sffamily
            \textbf{--- Sample B: Differential Phase Truncation ($m=4$) ---} \\
            
            \begin{quote}
                \ttfamily\footnotesize
                \textbf{\textcolor{keywordcolor}{INPUT CONTEXT} ($s_{1:3}$):} \\
                \normalfont\itshape\color{black}
                % [填入 Step 1-3，用省略号代替前文]
                "...[History]... This pattern is highly indicative of \underline{Disease Main}." \textcolor{gray}{(Model sees diagnosis hypothesis)}
                
                \vspace{0.2em}
                \ttfamily\footnotesize
                \textbf{\textcolor{keywordcolor}{TARGET LABEL} ($s_{4:L}, a$):} \\
                \normalfont\itshape\color{black}
                % [填入 Step 4 到最后]
                "We must distinguish this from \underline{Disease Alternative}. ... [Exclusion logic] ... confirmed as Disease Main."
            \end{quote}

            % ==========================================
            % SAMPLE 3: Late Stage (m=5)
            % ==========================================
            \vspace{0.2em}
            \normalfont\sffamily
            \textbf{--- Sample C: Conclusion Phase Truncation ($m=5$) ---} \\
            
            \begin{quote}
                \ttfamily\footnotesize
                \textbf{\textcolor{keywordcolor}{INPUT CONTEXT} ($s_{1:4}$):} \\
                \normalfont\itshape\color{black}
                % [填入 Step 1-4]
                "...Unlike Disease Main, Disease Alternative typically lacks \textbf{Feature X}..."
                
                \vspace{0.2em}
                \ttfamily\footnotesize
                \textbf{\textcolor{keywordcolor}{TARGET LABEL} ($s_{5}, a$):} \\
                \normalfont\itshape\color{black}
                % [填入 Step 5 结论]
                "Given the absence of Feature Z, the diagnosis is confirmed as \underline{Disease Main}."
            \end{quote}
            
            \vspace{0.2em}
        };
        
        % 标题 Tab
        \node[
            fill=framecolor,
            text=white,
            rounded corners={5pt},
            anchor=west,
            xshift=20pt
        ] at (box.north west) {
            \bfseries \sffamily \hspace{0.5em} Trajectory Augmentation Example III \hspace{0.5em}
        };
        
    \end{tikzpicture}
    
    \caption{\textbf{Augmented Samples for Case Study III.} The process starts with a complete Golden Reason Chain ($R$). The augmentation strategy creates multiple training instances by truncating $R$ at random steps $m$. The model is provided with the partial history (Input Context) and trained to recover the remaining logic (Target Label), effectively teaching it to handle both early-stage generation and late-stage differential deduction.}
\label{Augmentation2}
\end{figure}

\section{Implementation Details of PathReasoner-R1}
\label{sec:implementations}
\subsection{Baseline of PathReasoner-R1}
\label{subsec:baseline}
% 介绍slidechat / longnet，直接用slidechat里的内容
Drawing inspiration from SlideChat~\cite{SlideChat}, we approach the analysis of gigapixel Whole-Slide Images (WSIs) within a multimodal framework. We first tessellate the WSI into non-overlapping $224 \times 224$ patches, which are subsequently processed by a frozen, foundation patch-level encoder~\cite{CONCH} to extract fine-grained feature representations. To aggregate these patch embeddings, we employ LongNet~\cite{LongNet} as the slide encoder. By leveraging a sparse attention mechanism across the entire slide, LongNet captures both local nuances and global long-distance contextual relations. This process facilitates effective cross-spatial interaction and information propagation among patches, which is essential for capturing intricate morphological reasoning details.

The resulting slide-level features then serve as visual tokens for the vision-language model. Following the architecture of LLaVA, we utilize a projector layer to align the visual tokens with the textual embedding space. This alignment bridges the gap between general visual concepts and the specific pathology domain. Furthermore, through instruction fine-tuning, the model is endowed with the capability to output intermediate reasoning processes. This design enables the model to address complex medical queries ranging from diagnosis to detailed QA, thereby facilitating practical clinical deployment.

\subsection{WSI Pre-processing in PathReasoner-R1}
\label{subsec:preprocess}
Given the gigapixel scale of WSIs, following \cite{CLAM}, we first segment tissue regions from a WSI $X$ and partition them into a sequence of non-overlapping patches $\mathbf{X} = \{X_1, X_2, \dots, X_L\}$, where $X_i \in \mathbb{R}^{3 \times 224 \times 224}$ and $L$ denotes the sequence length. 
Subsequently, a pre-trained and frozen pathology vision encoder $E_{\text{patch}}$ \cite{CONCH} extracts patch-level features $\mathbf{x}=\{x_1, x_2, \dots, x_L\} \in \mathbb{R}^{L \times D}$, where $x_i = E_{\text{patch}}(X_i)$. 
To model global dependencies across the entire slide, we employ LongNet \cite{LongNet} as the slide-level encoder $E_{\text{slide}}$. LongNet utilizes a sparse attention mechanism (Dilated Attention) that achieves linear complexity, efficiently handling the extremely long sequences inherent in WSI data \cite{gigapath}. 
Finally, a projection layer $\sigma_{\text{proj}}$ aligns the features with the textual embedding space. The entire process is formulated as:
\begin{equation}
    \hat{\mathbf{x}} = \sigma_{\text{proj}}(E_{\text{slide}}(\mathbf{x})) \in \mathbb{R}^{L \times D_{\text{llm}}},
\end{equation}
where $D_{\text{llm}}$ is the dimension of the LLM embedding, and $\hat{\mathbf{x}}$ serves as the final visual representation.

\subsection{Training Configurations}
\label{subsec:training}
We implement our method using SlideChat \cite{SlideChat} as the backbone framework. All experiments are conducted on a cluster equipped with 8 $\times$ NVIDIA RTX 4090 48GB GPUs. We optimize the model using the AdamW \cite{AdamW} optimizer and employ Low-Rank Adaptation (LoRA) \cite{LoRA} for parameter-efficient fine-tuning.
Comprehensive hyperparameter settings are detailed in Table~\ref{tab:training_config}.
The training pipeline consists of two sequential stages:
\begin{itemize}
    \item \textbf{Stage 1: Reasoning SFT.} To elicit the model's reasoning capabilities, we first perform supervised fine-tuning (SFT) on a curated dataset comprising 200K Chain-of-Thought samples.
    \item \textbf{Stage 2: Reinforcement Learning.} Subsequently, we further optimize the model via reinforcement learning utilizing a 20K non-CoT dataset with a reduced learning rate to ensure training stability.
\end{itemize}

\begin{table}[htbp]
\centering
\caption{Hyperparameters and configurations for PathReasoner-SFT and PathReasoner-R1 training.}
\label{tab:training_config} % 给表格单独的 label
\renewcommand{\arraystretch}{1.2} 
\setlength{\tabcolsep}{10pt} % 稍微调整列宽适应页面

\begin{tabular}{lcc} 
\toprule
\textbf{Configuration} & \textbf{Stage-1 (PathReasoner-SFT)} & \textbf{Stage-2 (PathReasoner-R1)} \\
\midrule
\multicolumn{3}{l}{\textit{\textbf{Data \& Model}}} \\
\quad Dataset Type & CoT Dataset & Non-CoT Dataset \\
\quad Data Size & 200K\textsuperscript{*} & 20K \\
\quad Parameters & 7.79 B & 7.79 B \\
\quad LLM (Base+LoRA) Parameters & 7.77 B & 7.77 B \\
\quad LongNet Parameters & 4.21 M & 4.21 M \\
\quad Projector Parameters & 14.69 M & 14.69 M \\
\midrule
\multicolumn{3}{l}{\textit{\textbf{Training Hyperparameters}}} \\ 
\quad Batch Size & 1 & 1 \\
\quad Gradient Accumulation & 16 & 4 \\
\quad Learning Rate & $5 \times 10^{-5}$ & $1 \times 10^{-6}$ \\
\quad Epochs & 5 & 2  \\
\quad Optimizer & AdamW & AdamW \\
\quad Strategy & DeepSpeed ZeRO-3 & DeepSpeed ZeRO-3 \\
\quad Weight Decay & 0.05 & 0.0 \\
\quad Warmup Ratio & 0.03 & 0.00 \\
\quad Scheduler & Cosine & Cosine \\
\quad WSI Max Seq Length & 4,096 & 4,096 \\
\quad LoRA rank & 32 & 32 \\
\quad LoRA alpha & 64 & 64 \\
\quad G & - & 8 \\
\quad $\epsilon$ & - & 0.2 \\
\quad $\gamma$ & - & 0.03 \\
\quad GPU Configuration & $8$ & $8$ \\
\bottomrule
\end{tabular}
\begin{flushleft}
\footnotesize
\textit{Note:} \textsuperscript{*} indicates the dataset after trajectory sampling steps. 
\end{flushleft}
\end{table}

\section{Evaluation Metrics}
\label{sec:evaluation}

To comprehensively assess the quality of generated reports, we employ a multidimensional evaluation protocol that covers lexical overlap, semantic consistency, and clinical validity.
Specifically, we utilise standard Natural Language Generation (NLG) metrics, including BLEU \cite{BLEU} and ROUGE-1/2/L \cite{ROUGE}, to quantify n-gram overlap. To evaluate semantic alignment beyond surface-level textual matching, we incorporate BERTScore \cite{BERTScore}.
Acknowledging the complexity of pathology reporting, we further adopt an LLM-as-a-Judge approach utilising Qwen3-Max \cite{yang2025qwen3}.
Furthermore, the model's chain-of-thought is evaluated across two specialized dimensions using Qwen3-Max:
(i) Alignment Score (A-Score): Quantifies the factual consistency and semantic agreement between the generated reasoning trajectory and the ground-truth annotations.
(ii) Quality Score (Q-Score): Evaluates the intrinsic logical coherence, structural integrity, and clinical plausibility of the reasoning process itself, independent of the final answer.

\section{More experiments and ablation studies}
\label{sec:comparison}

\subsection{Prompts Used for Generation and Evaluation}
\label{subsec:prompts}
Figure \ref{fig:generation_prompt} illustrates the structured prompts designed to elicit multi-step histopathological reasoning within the PathReasoner framework. To evaluate the quality of VQA answers, we adopt an LLM-as-a-judge approach, with the corresponding scoring prompt presented in Figure \ref{fig:vqa_score}. Furthermore, we assess both the accuracy of the reasoning output relative to the ground truth (A-score) and the intrinsic quality of the reasoning process (Q-score). The specific prompts for these LLM-based evaluations are detailed in Figures \ref{fig:reasoning_Ascore} and \ref{fig:reasoning_Qscore}.

% 模型生成prompt
\begin{figure}[!p]
    \centering
    \begin{tikzpicture}
        % 主内容框
        \node[
            draw=framecolor,
            line width=1.2pt,
            rounded corners=8pt,
            inner sep=15pt,
            text width=0.9\textwidth, 
            align=left,
            font=\small,
            fill=codebg
        ] (box) {
            
            % --- Option 1 ---
            \textbf{\textsf{\color{framecolor}Prompt Option 1:}} \\
            You are an expert AI pathologist. Carefully analyze the provided whole slide image (WSI) to answer the following question. \\
            Follow these steps:
            \begin{itemize}[noitemsep, topsep=2pt, leftmargin=2.5em]
                \item[1.] Observe and describe key histopathological findings relevant to the question.
                \item[2.] Perform step-by-step clinical reasoning to connect findings with your conclusion.
                \item[3.] Provide the final concise answer.
            \end{itemize}
            Please respond in the exact format below: \\
            \texttt{<observe>} Histopathological Findings: ... \texttt{</observe>} \\
            \texttt{<think>} Clinical Reasoning: ... \texttt{</think>} \\
            \texttt{<answer>} Final Answer: ... \texttt{</answer>} \\[0.5em]
            \textbf{Question:} \texttt{\{Input\_question\}} 
            
            \vspace{0.8em}
            {\color{gray}\hrule height 0.4pt} % 这里使用 hrule 代替 draw，在 node 内部最稳定
            \vspace{0.8em}
            
            % --- Option 2 ---
            \textbf{\textsf{\color{framecolor}Prompt Option 2:}} \\
            You are an AI pathology assistant. Answer the following question based on the provided WSI. Please structure your reasoning clearly and respond in this exact format: \\
            \texttt{<observe>} Histopathological Findings: Describe key findings. \texttt{</observe>} \\
            \texttt{<think>} Clinical Reasoning: Explain your diagnostic reasoning step-by-step. \texttt{</think>} \\
            \texttt{<answer>} Final Answer: Provide the short, conclusive answer. \texttt{</answer>}\\[0.5em]
            \textbf{Question:} \texttt{\{Input\_question\}} 
            
            \vspace{0.8em}
            {\color{gray}\hrule height 0.4pt}
            \vspace{0.8em}
            
            % --- Option 3 ---
            \textbf{\textsf{\color{framecolor}Prompt Option 3:}} \\
            You are a digital pathology consultant. Analyze the provided WSI and answer the question using structured reasoning. Respond strictly in this format: \\
            \texttt{<observe>} ... \texttt{</observe>} \texttt{<think>} ... \texttt{</think>} \texttt{<answer>} ... \texttt{</answer>}\\[0.5em]
            \textbf{Question:} \texttt{\{Input\_question\}} 
            
        };
        
        % 标题 Tab
        \node[
            fill=framecolor,
            text=white,
            rounded corners=4pt,
            anchor=west,
            xshift=15pt,
            font=\sffamily\bfseries
        ] at (box.north west) {
            \hspace{0.5em}Prompts for PathReasoner Generation\hspace{0.5em}
        };
        
    \end{tikzpicture}
    \caption{Prompts used for to the generation of PathReasoner.}
    \label{fig:generation_prompt}
\label{ReasonPipeline8}
\end{figure}

% VQA打分prompt
\begin{figure}[!p]
    \centering
    \begin{tikzpicture}
        % 主内容框
        \node[
            draw=framecolor,
            line width=1.2pt,
            rounded corners=8pt,
            inner sep=15pt,
            text width=0.9\textwidth, 
            align=left,
            font=\small\rmfamily,
            fill=codebg
        ] (box) {
            
            % --- Section 1: Persona ---
            You are a senior pathologist with 20+ years of clinical experience. Your task is to score the model's answer to a medical visual question based on its clinical accuracy compared to the ground truth diagnosis.
            
            % --- Section 2: Scoring Criteria ---
            \textbf{\textsf{\color{framecolor}Scoring Criteria (1-5):}}
            \begin{itemize}[noitemsep, topsep=3pt, leftmargin=2.5em, labelsep=0.5em]
                \item[\textbf{5:}] \textbf{Perfectly correct.} Clinically equivalent to ground truth, uses precise terminology, no errors.
                \item[\textbf{4:}] \textbf{Mostly correct.} Minor phrasing issues (e.g., word order), but clinically sound and accurate.
                \item[\textbf{3:}] \textbf{Partially correct.} Captures key elements but misses critical details (e.g., tumor grade, margin status).
                \item[\textbf{2:}] \textbf{Related but incorrect.} Mentions a relevant category but gets the specific diagnosis wrong.
                \item[\textbf{1:}] \textbf{Incorrect or irrelevant.} Hallucinated, off-topic, or contradicts the ground truth.
            \end{itemize}
            
            % --- Section 3: Instructions ---
            \textbf{\textsf{\color{framecolor}Instructions:}}
            \begin{itemize}[noitemsep, topsep=3pt, leftmargin=1.5em]
                \item Focus on \textbf{clinical meaning}, not exact wording.
                \item Treat standard synonyms and abbreviations (e.g., ``IDC'') as acceptable.
                \item Penalize overgeneralization or inclusion of false information.
                \item Binary answers: incorrect responses must receive a low score (1 or 2).
            \end{itemize}
            
            % --- Section 4: Data & Final Constraint ---
            \textbf{Ground Truth:} \texttt{\{ground\_truth\}} \\
            \textbf{Model Prediction:} \texttt{\{model\_output\}} 
            
            \vspace{1em}
            \textbf{Constraint:} Respond \textbf{ONLY} with a single integer from 1 to 5.
        };
        
        % 标题 Tab
        \node[
            fill=framecolor,
            text=white,
            rounded corners=4pt,
            anchor=west,
            xshift=15pt,
            font=\sffamily\bfseries
        ] at (box.north west) {
            \hspace{0.5em}Prompts for Open-end VQA Evaluation\hspace{0.5em}
        };
        
    \end{tikzpicture}
    \caption{Prompts used to evaluate the quality of open-ended VQA answers.}
    \label{fig:vqa_score}
\end{figure}

% 推理质量打分prompt
\begin{figure}[!p]
    \centering
    \begin{tikzpicture}
        % 主内容框
        \node[
            draw=framecolor,
            line width=1.2pt,
            rounded corners=8pt,
            inner sep=15pt,
            text width=0.9\textwidth, 
            align=left,
            font=\small\rmfamily,
            fill=codebg
        ] (box) {
            
            % --- Role Definition ---
            You are an expert in computational pathology with extensive experience in histopathological analysis. Please evaluate the following AI-generated interpretation of a pathology image based on the \textbf{quality of its reasoning process}.
            
            \vspace{1.2em}
            
            % --- Assessment Criteria ---
            \textbf{\textsf{\color{framecolor}Assessment Criteria:}}
            \begin{itemize}[noitemsep, topsep=4pt, leftmargin=1.5em, labelsep=0.5em]
                \item \textbf{1. Logical Clarity:} Is the argument presented in a clear, stepwise manner? Are conclusions supported by prior statements without gaps or contradictions?
                \item \textbf{2. Evidence Alignment:} Does the reasoning explicitly connect each claim to observable morphological features (e.g., nuclear size, chromatin pattern, architecture)?
                \item \textbf{3. Professional Rigor:} Are pathological terms used precisely? Does the reasoning reflect sound principles and avoid speculative interpretations?
                \item \textbf{4. Explainability:} Would a practicing pathologist find the reasoning transparent? Does it articulate how visual findings lead to conclusions?
                \item \textbf{5. Comprehensiveness:} Does it address relevant diagnostic features and acknowledge key differential considerations or limitations?
            \end{itemize}
            
            \vspace{1.2em}
            
            % --- Scoring Scale ---
            \textbf{\textsf{\color{framecolor}Scoring Scale (for each dimension and overall):}}
            \begin{quote}
                5 = Excellent \quad 4 = Good \quad 3 = Fair \quad 2 = Poor \quad 1 = Very poor
            \end{quote}
            
            \vspace{0.8em}
            
            % --- Input Placeholder ---
            \textbf{AI-generated reasoning to evaluate:} \texttt{\{model\_reasoning\}}
        };
        
        % 标题 Tab
        \node[
            fill=framecolor,
            text=white,
            rounded corners=4pt,
            anchor=west,
            xshift=15pt,
            font=\sffamily\bfseries
        ] at (box.north west) {
            \hspace{0.5em}Prompts used to evaluate the quality of reasoning steps.\hspace{0.5em}
        };
        
    \end{tikzpicture}
    \caption{Prompts used to evaluate the quality of reasoning steps.}
    \label{fig:reasoning_Qscore}
\end{figure}

% 推理准确性打分prompt
\begin{figure}[!p]
\centering
\begin{tikzpicture}
% 主内容框
    \node[
    draw=framecolor,
    line width=1.2pt,
    rounded corners=8pt,
    inner sep=15pt,
    text width=0.9\textwidth,
    align=left,
    font=\small\rmfamily,
    fill=codebg
    ] (box) {

        % --- Persona and Task ---
        You are a senior pathologist with extensive experience in diagnostic reasoning. Below are two pieces of text:
        \begin{itemize}[noitemsep, topsep=3pt, leftmargin=1.5em]
            \item \textbf{Reference Reasoning:} The gold-standard explanation provided by an expert pathologist.
            \item \textbf{Model Reasoning:} The reasoning generated by an AI system analyzing the same pathology image.
        \end{itemize}
        
        \vspace{1em}
        \textbf{Task:} Your task is to score the Model Reasoning on a scale of 1 to 5 based on its \textbf{factual and logical alignment with the Reference Reasoning}. Focus on whether the model captures the same key observations, interpretive steps, and diagnostic logic.

        \vspace{1.2em}
        
        % --- Scoring Criteria ---
        \textbf{\textsf{\color{framecolor}Scoring Criteria:}}
        \begin{itemize}[noitemsep, topsep=4pt, leftmargin=2.5em, labelsep=0.5em]
            \item[\textbf{5:}] \textbf{Nearly identical.} Captures all critical findings and implications correctly in a similar reasoning flow.
            \item[\textbf{4:}] \textbf{Strong alignment.} Minor omissions or rephrasing, but no meaningful deviation in logic or facts.
            \item[\textbf{3:}] \textbf{Partial alignment.} Includes some correct elements but misses/misrepresents key diagnostic features.
            \item[\textbf{2:}] \textbf{Weak alignment.} Mentions related concepts but diverges significantly or omits essential evidence.
            \item[\textbf{1:}] \textbf{Minimal/No alignment.} Contains hallucinations, contradictions, or fails to reflect expert reasoning.
        \end{itemize}
        
        \vspace{1.2em}
        
        % --- Data Section ---
        \textbf{Reference Reasoning:} \texttt{\{reference\_reasoning\}} \\
        \textbf{Model Reasoning:} \texttt{\{model\_reasoning\}} 
        
        \vspace{1em}
        \centering
        \textbf{Respond ONLY with a single integer from 1 to 5.}
    };
    
    % 标题 Tab
    \node[
        fill=framecolor,
        text=white,
        rounded corners=4pt,
        anchor=west,
        xshift=15pt,
        font=\sffamily\bfseries
    ] at (box.north west) {
        \hspace{0.5em}Prompts used to evaluate the accuracy of reasoning steps\hspace{0.5em}
    };
    
    \end{tikzpicture}
    \caption{Prompts used to evaluate the accuracy of reasoning steps.}
    \label{fig:reasoning_Ascore}

\end{figure}

\subsection{Evaluation on Patch-Level Benchmarks}
\label{subsec:patch_level}
To ensure a comprehensive evaluation of the model's capabilities, we extended our experiments to the PathMMU benchmark. The complete testing set results are in Table~\ref{tab:patch}, and the results on the tiny testing set are in Table~\ref{table_pathmmu}. It is important to note that PathMMU consists primarily of patch-level images, presenting a significant granularity shift from the gigapixel WSIs that our PathReasoner is natively designed to process. Despite this structural discrepancy, our model demonstrates commendable robustness. On the PathMMU-test-tiny, although there remains a performance gap compared to models explicitly optimized for patch-level tasks (e.g., Patho-R1), PathReasoner-R1 maintains competitive performance and generalizes well to local visual details without specific patch-level fine-tuning. This indicates that the reasoning logic and diagnostic patterns acquired from global WSI contexts possess strong transferability, enabling the model to adapt effectively to diverse image scales and distinct data distributions.

\begin{table}[t] 
\caption{Comparison of vision-language models on PathMMU-test-tiny benchmark in terms of accuracy. \textbf{Bold}: best performance, \underline{underline}: second-best performance.}
\centering
\renewcommand{\arraystretch}{1.2}
\scriptsize 
\setlength{\tabcolsep}{10pt}
\begin{tabular}{lcccccc} % 列数从13改为7 (1+6)
\toprule \toprule
\multirow{2}{*}{\textbf{Method}} 
  & \multicolumn{6}{c}{\textbf{PathMMU-test-tiny (1139 samples)}} \\
\cmidrule(lr){2-7} 
  & Atlas & EduContent & PathCLS & PubMed & SocialPath & Overall \\
\midrule 
\rowcolor{green!10} %
Expert Performance        & 68.30 & 69.00 & 78.90 & 72.90 & 71.50 & 71.80 \\
\multicolumn{7}{l}{\cellcolor{gray!10}\textbf{Non-reasoning models}} \\ % colspan改为7
Qwen3-VL-8B-Instruct      & 25.00 & 21.57 & 7.34 & 24.56 & 22.48 & 20.90 \\
LLaVA-Med-7B & 31.25 & 21.18 & 13.56 & 31.32 & 18.35 & 23.79 \\
HuatuoGPT-Vision-7B       & 65.87 & 60.00 & 40.11 & 61.92 & 58.72 & 58.21 \\
MedGemma-4B-IT            & 33.15 & 29.10 & 8.05 & 28.80 & 31.10 & 26.65 \\
Quilt-LLaVA-7B       & 42.79 & 38.43 & 14.12 & 37.01 & 32.57 & 33.98 \\
SlideChat-7B              & 49.04 & 47.06 & 29.94 & 53.38 & 45.41 & 46.01 \\
WSI-LLaVA-7B              & 50.00 & 46.67 & 28.50 & 51.00 & 46.00 & 44.83 \\
\midrule \midrule
\multicolumn{7}{l}{\cellcolor{gray!10}\textbf{Models with reasoning ability}} \\ % colspan改为7
Qwen3-VL-8B-Thinking      & 44.23 & 49.41 & 24.86 & 44.84 & 40.83 & 41.88 \\
InternVL3.5-8B            & 58.17 & 54.90 & 42.94 & 57.65 & 60.55 & 55.40 \\
MedVLThinker-7B           & 49.04 & 47.06 & 29.94 & 53.38 & 45.41 & 46.01 \\
Patho-R1-7B               & \textbf{81.73} & \textbf{75.29} & {44.63} & \textbf{72.24} & \underline{67.89} & \textbf{69.53} \\
\rowcolor{blue!8}
{PathReasoner-SFT-7B} & 64.90 & 66.66 & \underline{45.19} & 65.83 & 63.40 & 61.19 \\
\rowcolor{blue!11}
{PathReasoner-R1-7B}  &\underline{73.55} & \underline{70.98} & \textbf{56.49} & \underline{69.39} & \textbf{68.08} & \underline{67.70} \\
\bottomrule \bottomrule
\end{tabular}
\vspace{-0.4em} 
\label{table_pathmmu}
\end{table}

\begin{figure}[t]
    \centering
    \includegraphics[width=\columnwidth]{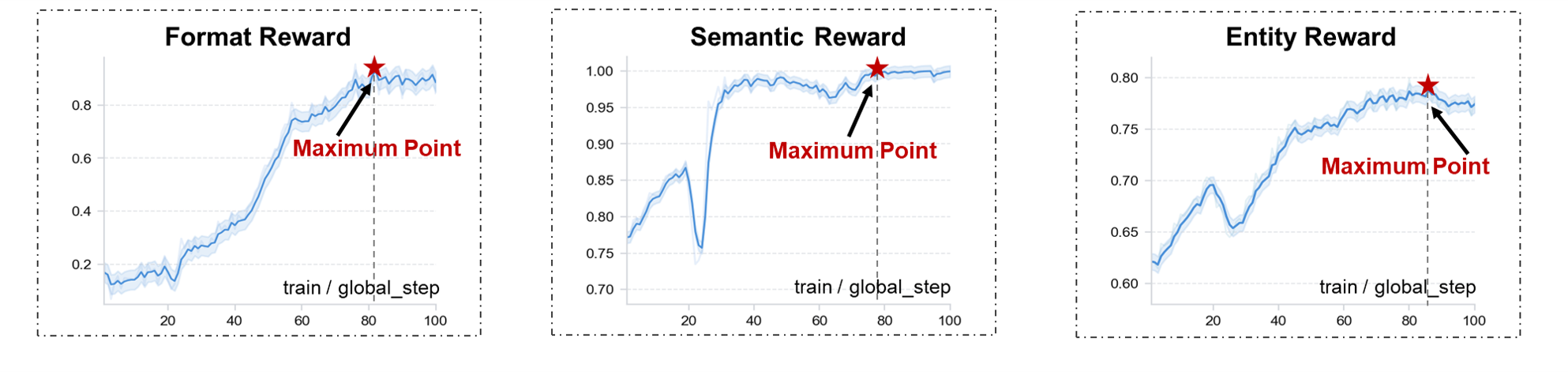}
    \caption{Training dynamics of the multi-granular reward components during the RL stage.}
    \label{fig:ablation_reward}
\end{figure}

\subsection{Evaluation on WSI-Level Benchmarks}
\label{subsec:wsi_level}
We conducted a comprehensive evaluation on the SlideBench-VQA-TCGA benchmark \cite{SlideChat} to assess model performance in whole-slide imaging (WSI) scenarios, with results detailed in Table~\ref{table3}. PathReasoner-R1 demonstrates superior performance, consistently outperforming other reasoning-integrated models (e.g., Patho-R1 \cite{Patho-R1} and InternVL3.5 \cite{InternVL}) across all metrics. Notably, the R1 variant shows a substantial improvement over its SFT baseline (PathReasoner-SFT), validating the efficacy of the reasoning strategies cultivated through reinforcement learning.

Although SlideChat maintains a competitive edge in specific visual recognition tasks—likely due to training distribution overlap with SlideBench-TCGA, PathReasoner-R1 surpasses it in complex diagnostic categories and achieves comparable accuracy in clinical questioning. It suggests that our model not only preserves robust visual perception for gigapixel slides but also excels in high-level diagnostic logic, effectively bridging the gap between raw feature extraction and sophisticated clinical reasoning.

\begin{table*}[t]
  \caption{Performance comparison on SlideBench-VQA-TCGA benchmark. Best results are \textbf{bolded}, second-best results are \underline{underlined}.}
  \centering
  \renewcommand{\arraystretch}{1.3}
  \scriptsize
  \setlength{\tabcolsep}{14pt}
  \begin{tabular}{l c cccc}
  \toprule \toprule
  \textbf{Model} & & \multicolumn{4}{c}{\textbf{SlideBench-VQA-TCGA}} \\
  \cmidrule(lr){3-6}
  & & Microscopy & Diagnosis & Clinical & Average \\
  \midrule
  \rowcolor{gray!9}
  \multicolumn{6}{l}{\textbf{Non-reasoning models}} \\
  Qwen2.5-VL-7B-Instruct  & T & 49.74 & 37.16 & 53.06 & 41.48 \\
  Qwen3-VL-8B-Instruct    & T & 57.85 & 39.38 & 69.39 & 46.16 \\
  LLaVA-Med-7B            & T & 35.60 & 21.05 & 42.86 & 26.27 \\
  HuatuoGPT-Vision-7B     & T & 58.64 & 39.58 & 60.20 & 45.89 \\
  MedGemma-4B-IT          & T & 49.48 & 36.96 & 57.14 & 41.55 \\
  Quilt-LLaVA-7B          & T & 44.76 & 20.24 & 52.04 & 28.72 \\
  SlideChat-7B            & S & \textbf{81.68} & \underline{73.21} & \textbf{72.45} & \textbf{75.36} \\
  WSI-LLaVA-7B            & S & 56.08 & 64.14 & 52.57 & 60.20 \\
  \midrule \midrule
  \rowcolor{gray!9}
  \multicolumn{6}{l}{\textbf{Models with reasoning ability}} \\
  Qwen3-VL-8B-Thinking    & T & 49.48 & 33.23 & 48.98 & 38.49 \\
  InternVL3.5-8B          & T & 56.28 & 45.52 & 68.37 & 49.83 \\
  MedVLThinker-7B         & T & 48.43 & 44.61 & 65.31 & 46.98 \\
  Patho-R1-7B             & T & 63.61 & 47.53 & 57.14 & 52.34 \\
  \rowcolor{blue!8}
  PathReasoner-SFT-7B & S & 75.73 & 65.05 & 67.35 & 67.98 \\
  \rowcolor{blue!11}
  PathReasoner-R1-7B & S & \underline{78.52} & \textbf{73.43} & \underline{72.39} & \underline{74.68} \\
  \bottomrule \bottomrule
  \end{tabular}
  \label{table3}
\end{table*}

\subsection{Training Dynamics Rewards Performance}
\label{subsec:rewards}
To analyze the stability of our optimization process, we tracked the trajectories of the three reward components—Format, Semantic, and Entity—throughout the RL training stage. As shown in Figure \ref{fig:ablation_reward}, the Format Reward shows a rapid, steady ascent, indicating the model's prioritization of learning the structured output format. Interestingly, we observe a distinct fluctuation in both Semantic and Entity Rewards, characterized by a temporary performance drop during the 20\%-30\% interval. This phenomenon suggests an adaptive adjustment phase where the model, in its effort to strictly align with the complex formatting rules (maximizing Format Reward), temporarily compromises its semantic coherence. However, as training proceeds beyond this exploration phase, the policy successfully harmonizes these objectives, with all rewards recovering and converging steadily to a plateau.

\subsection{Impact of Trajectory Augmentation}
\label{subsec:sampling} Table~\ref{table_sft_ablation} presents the ablation results regarding the trajectory augmentation strategy. We observe that training solely on full reasoning chains limits the model's potential. However, by introducing the random truncation mechanism, our method achieves a substantial improvement in overall accuracy. Notably, this gain is consistent across all sub-metrics (Microscopy, Diagnosis, and Clinical), demonstrating that the augmented data effectively regularizes the model and prevents overfitting to specific path templates.

\begin{table}[t]
\caption{Ablation study on the impact of trajectory augmentation strategies. By truncating reasoning chains at intermediate steps, we scale the training data and enhance the robustness of autoregressive logic.}
\centering
\renewcommand{\arraystretch}{1.20}
\setlength{\tabcolsep}{8pt} 
\small
\begin{tabular}{lccccc}
\toprule
\toprule
\multicolumn{2}{c}{Setting} & \multicolumn{4}{c}{SlideBench-TCGA} \\
\cmidrule(lr){1-2} \cmidrule(lr){3-6}
Strategy & Data Size & Microscopy & Diagnosis & Clinical & Average \\
\midrule
Full Sequence Only & $\sim$20K & 70.15 & 61.20 & 63.45 & 63.67 \\
Random Masking & $\sim$200K & 72.40 & 63.15 & 65.20 & 65.69 \\
\textbf{Trajectory Augments} & \textbf{$\sim$200K} & \textbf{78.52} & \textbf{73.43} & \textbf{72.39} & \textbf{74.68} \\
\bottomrule
\bottomrule
\end{tabular}
\label{table_sft_ablation}
\vspace{-18pt}
\end{table}

\subsection{Qualitative Results}
\label{subsec:qualitative}
\begin{figure*}[t]
  \centering
  \includegraphics[width=\textwidth]{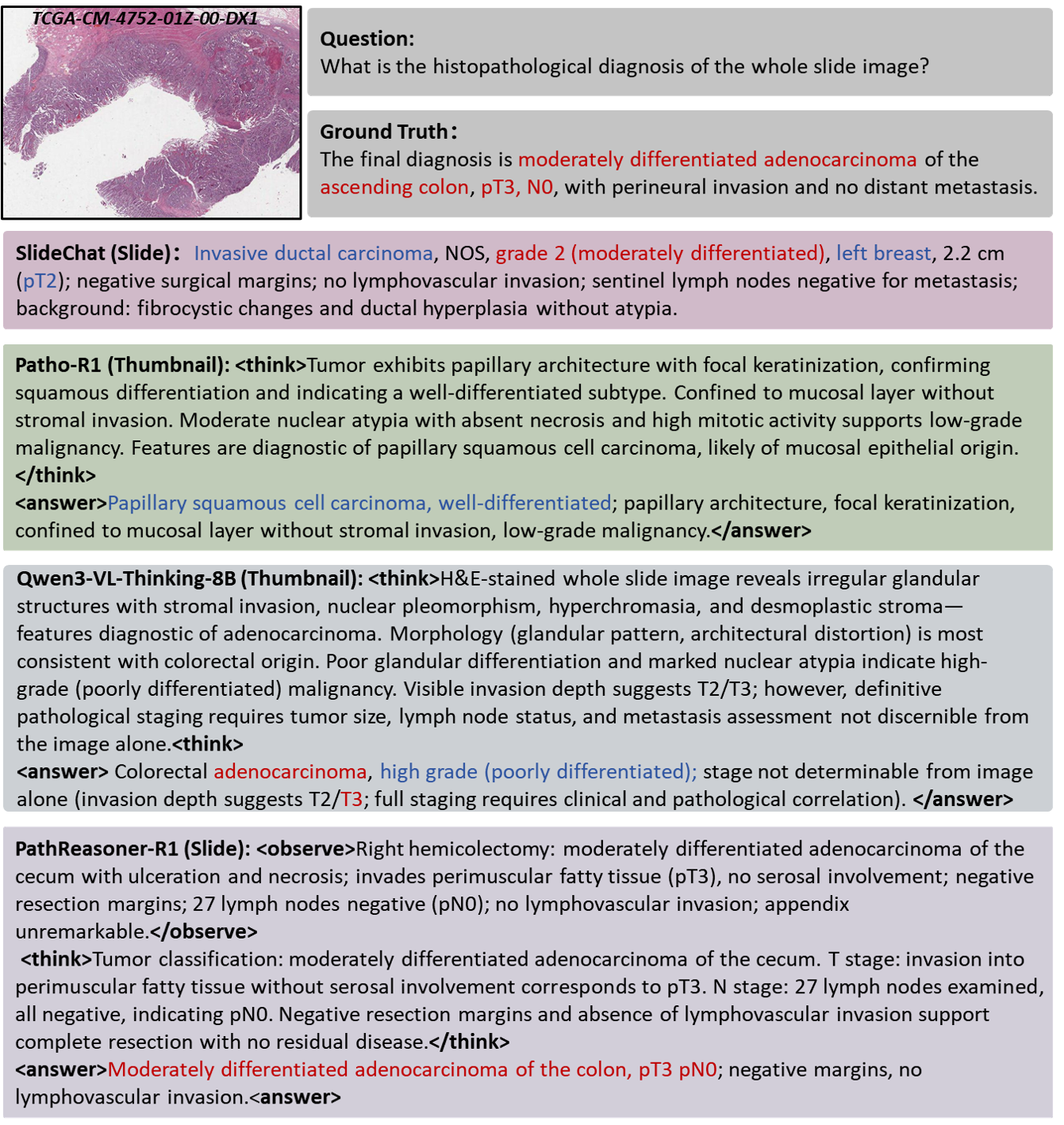}
  \caption{Qualitative comparison of model reasoning on TCGA Sample 1. We visualize the diagnostic outputs from PathReasoner-R1 and state-of-the-art CPath VLMs. Correct predictions are highlighted in red, while incorrect ones are in blue.}
  \label{fig:Qualitative_sample2}
\end{figure*}

\begin{figure*}[t]
  \centering
  \includegraphics[width=\textwidth]{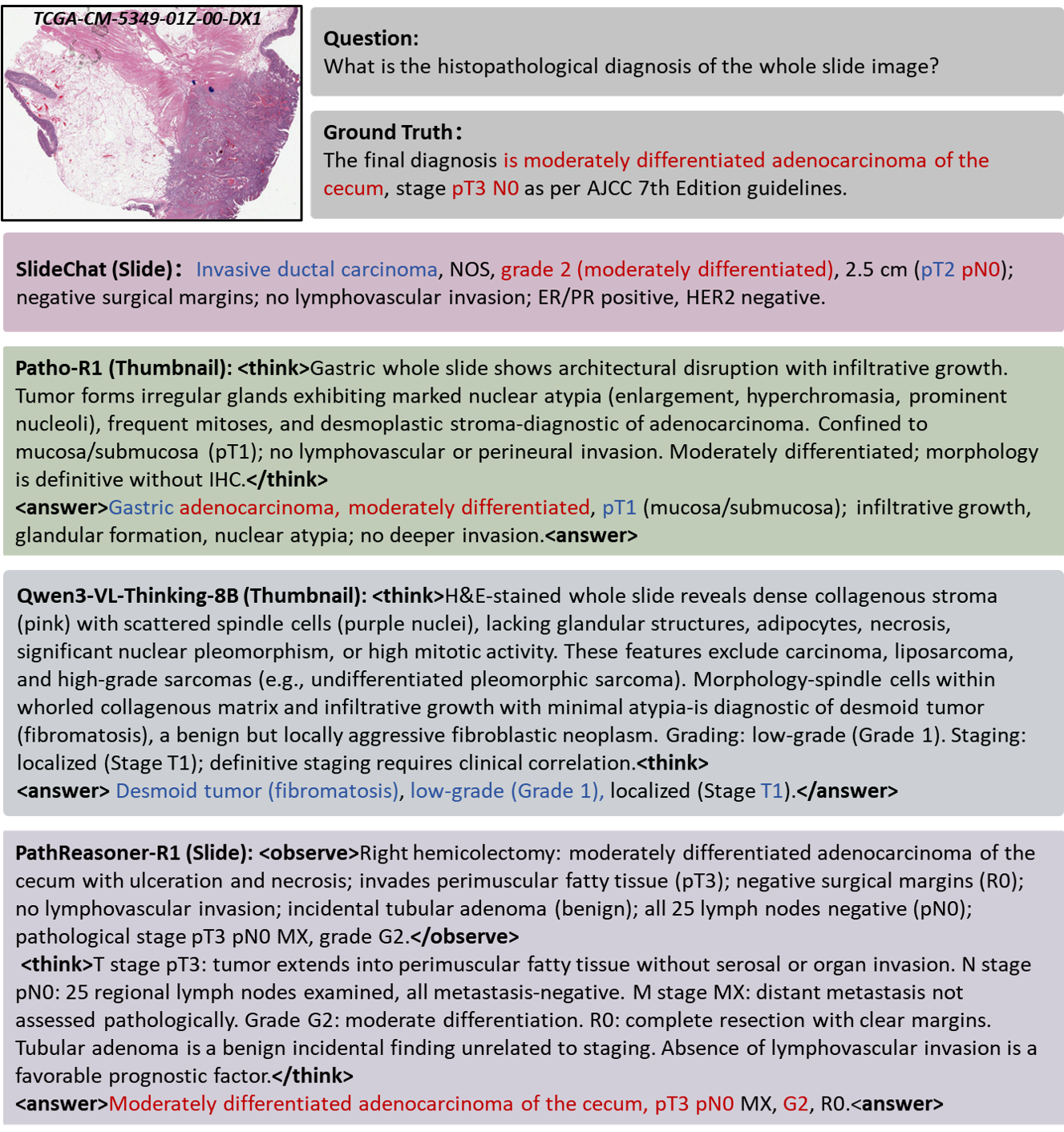}
  \caption{Qualitative comparison of model reasoning on TCGA Sample 2. We visualize the diagnostic outputs from PathReasoner-R1 and state-of-the-art CPath VLMs. Correct predictions are highlighted in red, while incorrect ones are in blue.}
  \label{fig:Qualitative_sample3}
\end{figure*}

In this part, we present additional qualitative comparison results across PathReasoner-R1 and the state-of-the-art CPath VLMs, as shown in Figures \ref{fig:Qualitative_sample2} and \ref{fig:Qualitative_sample3}. Predictions aligning with ground truth answers are highlighted in red, while mismatches are marked in blue. The results demonstrate that our PathReasoner-R1 achieves superior accuracy through an interpretable, step-wise reasoning process that explicitly enhances model transparency and reliability.

%%%%%%%%%%%%%%%%%%%%%%%%%%%%%%%%%%%%%%%%%%%%%%%%%%%%%%%%%%%%%%%%%%%%%%%%%%%%%%%
%%%%%%%%%%%%%%%%%%%%%%%%%%%%%%%%%%%%%%%%%%%%%%%%%%%%%%%%%%%%%%%%%%%%%%%%%%%%%%%

\end{document}